\documentclass[11pt]{article}
	
\newcommand{\blind}{0}

\makeatletter
\renewcommand\section{\@startsection {section}{1}{\z@}%
   {-3.5ex \@plus -1ex \@minus -.2ex}%
   {2.3ex \@plus.2ex}%
   {\normalfont\fontfamily{phv}\fontsize{16}{19}\bfseries}}
\renewcommand\subsection{\@startsection{subsection}{2}{\z@}%
 {-3.25ex\@plus -1ex \@minus -.2ex}%
 {1.5ex \@plus .2ex}%
 {\normalfont\fontfamily{phv}\fontsize{14}{17}\bfseries}}
\renewcommand\subsubsection{\@startsection{subsubsection}{3}{\z@}%
{-3.25ex\@plus -1ex \@minus -.2ex}%
 {1.5ex \@plus .2ex}%
 {\normalfont\normalsize\fontfamily{phv}\fontsize{14}{17}\selectfont}}
\makeatother

\usepackage{amsmath}
\usepackage{graphicx}
\usepackage{enumerate}
\usepackage{natbib} 
\usepackage{url} 
\usepackage[table,dvipsnames]{xcolor}
\usepackage{colortbl}

\usepackage{booktabs} 
\usepackage{amsfonts}
\usepackage{nicefrac}
\usepackage{microtype}
\usepackage{graphicx}
\usepackage{amsmath, amssymb,amsthm}
\usepackage{mathtools}
\usepackage{algorithm,algorithmic,setspace}

\usepackage{esvect}
\usepackage{bm}
\usepackage{subcaption}

\usepackage{multirow}
\usepackage{geometry, array, tabularray}
\usepackage{ctable} 

\usepackage{hyperref}
\hypersetup{
    citecolor = Green,
    colorlinks = true,
    linkcolor = blue
    }
\usepackage[
        nameinlink,
    ]{cleveref}

\usepackage{eqparbox}

\newcommand{\bfzero}{\mathbf 0}
\newcommand{\bfA}{\mathbf A}

\newcommand{\bfc}{\mathbf c}
\newcommand{\bfe}{\mathbf e}

\newcommand{\bff}{\mathbf f}

\newcommand{\bfI}{\mathbf I}

\newcommand{\bfK}{\mathbf K}
\newcommand{\bfL}{\mathbf L}

\newcommand{\bfM}{\mathbf M}
\newcommand{\bfm}{\mathbf m}

\newcommand{\bfP}{\mathbf P}
\newcommand{\bfQ}{\mathbf Q}
\newcommand{\bfR}{\mathbf R}
\newcommand{\bfS}{\mathbf S}
\newcommand{\bfs}{\mathbf s}

\newcommand{\bfu}{\mathbf u}

\newcommand{\bfX}{\mathbf X}
\newcommand{\bfx}{\mathbf x}
\newcommand{\bfy}{\mathbf y}
\newcommand{\bfz}{\mathbf z}
\newcommand{\bfZ}{\mathbf Z}
\newcommand{\bfalpha}{\boldsymbol{\alpha}}
\newcommand{\bfbeta}{\boldsymbol{\beta}}
\newcommand{\bfphi}{\boldsymbol{\phi}}

\newcommand{\bfmu}{\boldsymbol{\mu}}
\newcommand{\bftheta}{\boldsymbol{\theta}}

\newcommand{\calA}{\mathcal A}

\newcommand{\calD}{\mathcal D}

\newcommand{\calH}{\mathcal H}
\newcommand{\calGP}{\mathcal{GP}}
\newcommand{\calL}{\mathcal L}
\newcommand{\calM}{\mathcal M}
\newcommand{\calN}{\mathcal N}
\newcommand{\calO}{\mathcal O}
\newcommand{\calP}{\mathcal P}
\newcommand{\calS}{\mathcal S}

\newcommand{\bbE}{\mathbb E}

\newcommand{\bbI}{\mathbb I}
\newcommand{\bbN}{\mathbb N}

\newcommand{\bbR}{\mathbb R}
\newcommand{\B}{\mathrm{B}}
\newcommand{\Var}{\mathrm{Var}}

\newcommand{\cov}{\mathrm{cov}}

\newcommand{\hatq}{\hat{q}}

\newcommand{\hatmu}{\hat{\mu}}
\newcommand{\hatk}{\hat{k}}

\newcommand{\underh}{\underline{h}}
\newcommand{\underk}{\underline{k}}
\newcommand{\underl}{\underline{l}}

\definecolor{golden}{rgb}{0.9, 0.94, 0.0}

\begin{document}
\def\spacingset#1{\renewcommand{\baselinestretch}%
    {#1}\small\normalsize} \spacingset{1}

\if0\blind{
    \title{\bf Aggregation Models with Optimal Weights for Distributed Gaussian Processes}
    \author{Haoyuan Chen $^a$ and Rui Tuo $^a$\thanks{Corresponding author: \texttt{ruituo@tamu.edu}} \\
    $^a$ Department of Industrial and Systems Engineering\\ Texas A\&M University\\ College Station, TX 77843, USA
    }
    \date{}
    \maketitle
} \fi

\if1\blind{
    \title{\bf Aggregation Models with Optimal Weights for Distributed Gaussian Processes}
    \author{Author information is purposely removed for double-blind review
}
   
\bigskip
\bigskip
\bigskip
\begin{center}
    {\LARGE\bf Aggregation Models with Optimal Weights for Distributed Gaussian Processes}
\end{center}
\medskip
} \fi
\bigskip
		
\begin{abstract}
GP models have received increasing attention in recent years due to their superb prediction accuracy and modeling flexibility. To address the computational burdens of GP models for large-scale datasets, distributed learning for GPs are often adopted. Current aggregation models for distributed GPs is not time-efficient when incorporating correlations between GP experts. In this work, we propose a novel approach for aggregated prediction in distributed GPs. The technique is suitable for both the exact and sparse variational GPs. The proposed method incorporates correlations among experts, leading to better prediction accuracy with manageable computational requirements. As demonstrated by empirical studies, the proposed approach results in more stable predictions in less time than state-of-the-art consistent aggregation models.
\end{abstract}
			
\noindent%

\spacingset{1.5} 

\section{Introduction}
\textit{Gaussian processes} (GPs) \citep{rasmussen2003gaussian} are a powerful tool for modeling and inference in various areas of machine learning, such as regression \citep{o1978curve,bishop1995neural}, classification \citep{kuss2005assessing}, signal processing \citep{liutkus2011gaussian}, and robotics and control \citep{deisenroth2013gaussian}. Despite their strengths, GPs face significant challenges when applied to large-scale datasets due to the computational burden of inverting large covariance matrices.

To address these challenges, numerous efficient algorithms have been proposed. Sparse approximations using $m$ inducing points \citep{titsias2009variational} can reduce the computational complexity from $\calO(n^3)$ to $\calO(m^2 n)$ for a dataset of size $n$. \citet{wilson2015kernel} employed structured kernel interpolation combined with Kronecker and Toeplitz algebra to handle a large number of inducing points. \citet{hartikainen2010kalman} reformulated GP regression as linear-Gaussian state space models, solving them with classical Kalman filtering theory. \citet{katzfuss2021general} presented a general
Vecchia framework for GP predictions linear computational complexity in the total number of datasets. \citet{chen2022kernel} represented the kernels via a sparse linear transformation to help reduce complexity. More recently, \citet{lin2023sampling,lin2023stochastic} showed that a stochastic dual descent algorithm can efficiently yield competitive GP predictions. However, those approximation methods require at least $\calO(n)$ time, which makes them impractical for very large $n$.

Another direction to address the computational issue is distributed learning, which involves distributing computations across multiple units, often referred to as nodes or experts. In distributed learning \citep{dean2012large}, the dataset is partitioned and processed in parallel, with results aggregated to form the final model. By applying the previously mentioned GP approximation methods in distributed learning setup across $M$ units, the time complexity can be reduced to $\calO(n /M)$. Recent aggregation models for distributed GPs \citep{deisenroth2015distributed} rely on independence assumptions and cannot offer \textit{consistent} predictions, meaning the aggregated predictive distribution does not converge to the true underlying predictive distribution as the training size $n$ increases to infinity. To overcome this inconsistency,  \citet{rulliere2018nested} introduced the \textit{nested point-wise aggregation of experts} (NPAE), which incorporates covariances between the experts' predictions but is very time-consuming. \citet{liu2018generalized} proposed \textit{generalized robust Bayesian committee machine} (grBCM), an efficient and consistent algorithm for distributed GPs.

In this work, we introduce a novel aggregation model for distributed GP to improve time complexity. Our methodology is based on GP regressions using the \textit{optimized combination technique} (OptiCom) introduced in \Cref{sec:GP with OptiCom}. The proposed method considers the correlations between the experts to improve the consistency of the aggregated predictions. It is also less time-consuming than NPAE and grBCM when the number of GP experts $M$ is not very large. Empirical studies show that the proposed algorithm outperforms state-of-the-art aggregation models in computational efficiency while providing stable and consistent predictions.

The remainder of this paper is structured as follows: \Cref{sec:back} reviews the background and related work on GPs and distributed GPs. \Cref{sec:GP with OptiCom} presents the algorithm for extending OptiCom to GP regressions. \Cref{sec:DistGP with optimal weights} details our proposed algorithm for distributed GPs. \Cref{sec:exp} presents experimental results and comparisons. Finally, \Cref{sec:conc} concludes the paper with discussions.

\section{Background}
\label{sec:back}
In this section, we provide the necessary background for understanding the proposed methodology. We begin with a review of Gaussian processes in \Cref{subsec:GPs}, covering standard GP regression in \Cref{subsubsec: gp reg}, training procedures in \Cref{subsubsec: gp train}, and sparse variational GP approximations in \Cref{subsubsec:SVGP}. We then introduce distributed GPs in \Cref{subsec:distributed GP}, covering both training strategies in \Cref{subsubsec:Distributed GP training} and existing aggregation methods in \Cref{subsubsec: aggreg pred}.

\subsection{GPs}\label{subsec:GPs}
\subsubsection{GP regression}\label{subsubsec: gp reg}
A \textit{Gaussian process} (GP) is a collection of random variables, any finite number of which has a multivariate normal distribution. It is defined by a mean function $\mu(\cdot)$ and a kernel function $k(\cdot,\cdot)$: $f(\cdot) \sim \mathcal{GP}(\mu(\cdot),k(\cdot,\cdot))$. 

Suppose we have a set of training points $\bfX=\{\bfx_i\}_{i=1}^n$ and the observations $\bfy=(y_1,\ldots,y_n)^{\top}$ where $y_i=f(\bfx_i)+\epsilon_i$ with the i.i.d. noise $\epsilon_i \sim \calN(0,\sigma_{\epsilon}^2)$, $f:\bbR^d \rightarrow \bbR$ is a latent function. The GP regression imposes a GP prior over the latent function, and we have $\bff = f(\bfX) \sim \calN(\bfmu_{\bff}, \bfK_{\bff \bff})$ where 
$\bfmu_{\bff}= \big(\mu(\bfx_1),\cdots,\mu(\bfx_n)\big)^{\top}$ and $\bfK_{\bff \bff}=\big[k(\bfx_i,\bfx_j)\big]_{i,j=1}^n$.  
In this work, we assume $f$ is a zero-mean GP, i.e., $\mu(\cdot)=0$, the posterior distribution $p(f \vert \bfy)=\calN \big( \mu_{f \vert \bfy}, k_{f \vert \bfy} \big)$ takes the form\begin{subequations}\label{eq:conditional-gp}
    \begin{equation}
    \mu_{f \vert \bfy}(\cdot)= \bfK_{(\cdot)\bff} \widetilde{\bfK}_{\bff \bff} ^{-1}\bfy,\label{eq:conditional-mean}
    \end{equation}
    \begin{equation}
     k_{f \vert \bfy}(\cdot, \cdot') = k(\cdot, \cdot') - \bfK_{(\cdot)\bff} \widetilde{\bfK}_{\bff \bff}^{-1} \bfK_{\bff (\cdot')}, \label{eq:conditional-variance}
     \end{equation}
\end{subequations}
where $\widetilde{\bfK}_{\bff \bff}= \bfK_{\bff \bff} + \sigma_{\epsilon}^2 \bfI_n$ is the covariance matrix of all training points $\bfX$ with diagonal observation noise, $\sigma_{\epsilon}^2>0$ is the noise variance, $\bfI_n$ is a $n\times n$ identity matrix,  $\bfK_{(\cdot) \bff} = k(\cdot,\bfX)=\big[ k(\cdot,\bfx_i) \big]_{i=1}^{i=n}=k(\bfX,\cdot)^{\top}=\bfK_{\bff(\cdot)}^{\top}$ is the cross-covariance matrix.

\subsubsection{GP training}
\label{subsubsec: gp train}
The hyperparameters $\bftheta$ of GPs, which may include GP variance, kernel lengthscale, are commonly learned by maximizing the \textit{log-marginal likelihood} 
$\calL(\bftheta) = \log p(\bfy \vert \bff; \bftheta)$.
A standard way of obtaining the optimized hyperparameters is by taking the derivative of each $\theta \in \bftheta$. 

\subsubsection{Sparse variational GP (SVGP) }
\label{subsubsec:SVGP}
Exact GP regression and training as described in \Cref{subsubsec: gp reg} and \Cref{subsubsec: gp train} require $\calO(n^3)$ time due to the computation of the terms $\widetilde{\bfK}_{\bff \bff}^{-1}$, $\log\det (\widetilde{\bfK}_{\bff \bff}^{-1})$, and $\text{tr}\Big( \widetilde{\bfK}_{\bff \bff}^{-1} \frac{\partial \widetilde{\bfK}_{\bff \bff}}{\partial \theta} \Big)$, which is prohibitive when $n$ is large. To solve this problem, \citet{titsias2009variational} introduced \textit{sparse variational Gaussian process} (SVGP), a variational framework for sparse GPs using $m \ll n$ \textit{inducing variables} $\bfu=(u_1,\ldots,u_m)^{\top}$ with $m$ associated \textit{inducing inputs} $\bfZ=\{\bfz_i\}_{i=1}^m$, $\bfz_i \in \bbR^{d}$, which reduces the time complexity to $\calO(nm^2)$. The variational distribution $q(\bfu) = \calN(\bfm_{\bfu}, \bfS_{\bfu\bfu})$ yields the predictive distribution $\hatq(f)=\calN(\hatmu_{f}^{\bfu}, \hatk_{f}^{\bfu})$ as follows: 
\begin{subequations}\label{eq:svgp-optimal-conditional-gp}
    \begin{equation}
    \hat{\mu}_{f}^{\bfu}(\cdot)= \sigma_{\epsilon}^{-2} \bfK_{(\cdot)\bfu} \bfM_{\bfu \bfu}^{-1} \bfK_{\bfu \bff} \bfy,\label{eq:svgp-optimal-conditional-mean}
    \end{equation}
    \begin{equation}
     \hat{k}_{f}^{\bfu}(\cdot, \cdot') = k(\cdot, \cdot') - \bfK_{(\cdot)\bfu} (\bfK_{\bfu \bfu}^{-1} - \bfM_{\bfu \bfu}^{-1}) \bfK_{\bfu (\cdot')}, \label{eq:svgp-optimal-conditional-variance}
     \end{equation}
\end{subequations}
where $\bfK_{ (\cdot) \bfu }=\big[k(\cdot,\bfz_i)\big]_{i=1}^{m} = \bfK_{ \bfu (\cdot) }^{\top}$, 
$\bfK_{\bfu \bff} = \big[ k(\bfz_{i}, \bfx_{j}) \big]_{i,j=1}^{m,n} = \bfK_{\bff \bfu}^{\top}$, and $\bfM_{\bfu \bfu} = \big[ \bfK_{\bfu \bfu} + \sigma_{\epsilon}^{-2} \bfK_{\bfu \bff} \bfK_{\bff \bfu}\big]$.

The SVGP framework introduced here is essential for our distributed learning methodology in \Cref{sec:DistGP with optimal weights}. By employing SVGP approximations at the local expert level before aggregation, we achieve a double reduction in computational cost: first, inducing points approximation reduces the cost from $\calO(n^3)$ to $\calO(nm^2)$; second, data parallelization across $M$ experts further reduces the time complexity to $\calO(nm^2/M)$.

\subsection{Distributed GP}\label{subsec:distributed GP}
While SVGP reduces complexity from $\calO(n^3)$ to $\calO(nm^2)$, this can still be impractical when $n$ is extremely large such as  billions of observations. Distributed learning provides a natural solution by partitioning the data across $M$ computational units (experts) and processing them in parallel, further reducing per-unit complexity to $O(n m^2 / M)$.

This section introduces the distributed GP framework that motivates our main contributions. \Cref{subsubsec:Distributed GP training} describes how local experts are trained independently on partitioned local datasets. \Cref{subsubsec: aggreg pred} reviews existing aggregation methods for distributed GPs and highlights their strengths and limitations.

\subsubsection{Distributed GP training}
\label{subsubsec:Distributed GP training}
Distributed GPs \citep{deisenroth2015distributed} enable the parallel training of GPs across different experts to accelerate the computing process for large-scale datasets. Note that distributed GPs are not inherently private because they involve data sharing or centralized aggregation, whereas federated GPs preserve privacy by keeping raw data local and only exchanging model updates. Assuming the same model as in \Cref{subsubsec: gp reg}, we first partition the global training dataset $\calD :=\{ \bfX,\bfy \}$ of size $n$ into $M$ local datasets $\calD_i=\{ \bfX_i,\bfy_i \}$, each of size $n_i$ for $i=1,\ldots,M$. Each local dataset $\calD_i$ corresponds to an expert $\calM_i$. Clearly, $n=\sum_{i=1}^{M} n_i$. We suppose that all local datasets share the same hyperparameters $\bftheta$. The global objective for distributed GP training is to maximize the global marginal log-likelihood $\log p(\bfy \vert \bff, \bftheta)$, where $\bff = f(\bfX)$. Assuming independence among all the experts $\{\calM_i\}_{i=1}^{M}$, the global marginal log-likelihood is approximated as:  
\begin{align}\label{eq:fact-gp train}
    \log p(\bfy \vert \bff; \bftheta) &\approx \sum_{i=1}^{M} \log p_i(\bfy_i \vert \bff_i; \bftheta) 
    = -\frac{1}{2} \sum_{i=1}^{M} \Big( \bfy_i^{\top} \widetilde{\bfK}_{\bff_i \bff_i}^{-1} \bfy_i + \log\det(\widetilde{\bfK}_{\bff_i \bff_i})  + n \log(2\pi) \Big)
\end{align}
where $p_i(\bfy_i \vert \bff_i, \bftheta) \sim \calN(\bfzero, \widetilde{\bfK}_{\bff_i \bff_i})$ is the local marginal likelihood of the $i$-th model $\calM_i$ with $\widetilde{\bfK}_{\bff_i \bff_i} = \bfK_{\bff_i \bff_i} + \sigma_{\epsilon}^2 \bfI_{n_i}$ and $\bfK_{\bff_i \bff_i} = k(\bfX_i,\bfX_i) \in \bbR^{n_i \times n_i}$, $\bff_i = f(\bfX_i) \in \bbR^{n_i}$.

The factorized approximation of the log-likelihood in \cref{eq:fact-gp train} (FACT) approximates the full covariance matrix $\bfK_{\bff \bff} = k(\bfX, \bfX)$ by a diagonal block matrix $\text{diag}[\bfK_{\bff_1,\bff_1},\ldots,\bfK_{\bff_M,\bff_M}]$, reducing the time complexity to $\calO(n_i^3)$. Other optimization techniques include Federated Averaging (FedAvg) \citep{mcmahan2017communication}, which averages local model updates; FedProx \citep{li2020federated}, which improves FedAvg by addressing client heterogeneity; ADMM \citep{boyd2011distributed}, which solves a consensus optimization problem by introducing auxiliary variables and dual updates.

\subsubsection{Aggregated prediction}
\label{subsubsec: aggreg pred}
After training local experts, we aggregate local predictions to form a global prediction. Let $\mu_i(\cdot)$ and $\sigma_i^2(\cdot)$ denote the predictive mean and variance from expert $\calM_i$. Common aggregation methods for GP experts include \textit{product-of-experts} (PoE) \citep{hinton2002training}, \textit{generalised product-of-experts} (gPoE) \citep{cao2014generalized}, \textit{Bayesian committee machine} (BCM) \citep{tresp2000bayesian}, \textit{robust Bayesian committee machine} (rBCM) \citep{deisenroth2015distributed}, \textit{generalized robust Bayesian committee machine} (grBCM) \citep{liu2018generalized}, and \textit{nested pointwise aggregation of
experts} (NPAE) \citep{rulliere2018nested}. PoE, gPoE, BCM and rBCM produce a global prediction using a weighted average of local precisions, the joint mean $\mu_{\calA}$ and joint precision $\sigma_{\calA}^{-2}$ are given by
\begin{subequations}\label{eq:poe and bcm}
    \begin{equation}
        \mu_{\calA}(\cdot) = \sigma_{\calA}^2 (\cdot) \sum_{i=1}^{M} \beta_i \sigma_i^{-2} (\cdot) \mu_i(\cdot),
    \end{equation}
    \begin{equation}
        \sigma_{\calA}^{-2}(\cdot) = \sum_{i=1}^{M} \beta_i \sigma_{i}^{-2}(\cdot) + (1 - \sum_{i=1}^M \beta_i) \sigma_{**}^{-2}(\cdot),
    \end{equation}
\end{subequations}
where $\sigma_{**}^{-2}(\cdot)=k(\cdot,\cdot) + \sigma_{\epsilon}^2$ is the prior variance, serving as a correction term for the BCM family. The weights, $\beta_i$, are defined as follows: $\beta_i=1$ for PoE and BCM, and $\beta_i= 0.5 (\log\sigma_{**}^2(\cdot) - \log\sigma_i^2(\cdot))$ for gPoE and rBCM.

For grBCM, the $M$ experts are divided into two groups: the global expert $\calM_c$, trained on dataset $\calD_c = \calD_1$ of size $n_c$, and the local experts $\{\calM_i\}_{i=2}^{M}$, trained on datasets $\{\calD_i\}_{i=2}^M$. The augmented dataset $\calD_{+i}=\{ \calD_c, \calD_i \}$ leads to a new expert $\calM_{+i}$ for $i=2,\ldots,M$. The joint mean and joint precision are given by:
\begin{subequations}\label{eq:grbcm}
    \begin{equation}
        \mu_{\calA}(\cdot) = \sigma_{\calA}^2(\cdot) \left(  \sum_{i=2}^{M} \beta_i \sigma_{+i}^{-2} (\cdot) \mu_{+i}(\cdot) - \Big(\sum_{i=2}^M \beta_i - 1 \Big) \sigma_{c}^{-2}(\cdot)\mu_{c}(\cdot) \right),
    \end{equation}
    \begin{equation}
        \sigma_{\calA}^{-2}(\cdot) = \sum_{i=2}^{M} \beta_i \sigma_{+i}^{-2}(\cdot) - \Big( \sum_{i=2}^M \beta_i - 1 \Big) \sigma_{c}^{-2}(\cdot),
    \end{equation}
\end{subequations}
where $\mu_c$ and $\sigma_c^{-2}$ are the mean and precision of expert $\calM_c$, and $\mu_{+i}$ and $\sigma_{+i}^{-2}$ are the mean and precision of expert $\calM_{+i}$. The weights are defined as $\beta_2=1$, and $\beta_i=0.5 (\log\sigma_{c}^2(\cdot) - \log\sigma_{+i}^2(\cdot))$ for $i=3,\ldots,M$. 

NPAE leverages the covariances between experts to ensure consistent predictions. The joint mean and joint variance are computed as follows:
\begin{subequations}\label{eq:npae}
    \begin{equation}
        \mu_{\calA}(\cdot) = \bfK_{(\cdot)\calA} \bfK_{\calA \calA}^{-1} \bfmu_{\calA},
    \end{equation}
    \begin{equation}
        \sigma_{\calA}^2(\cdot) = k(\cdot,\cdot) - \bfK_{(\cdot)\calA} \bfK_{\calA \calA}^{-1} \bfK_{\calA(\cdot)} +\sigma_{\epsilon}^2,
    \end{equation}
\end{subequations}
where $\bfK_{(\cdot) \bff_i}=k(\cdot, \bfX_i) = \bfK_{\bff_i (\cdot)}^{\top}$, $\bfK_{\bff_i \bff_j}=k(\bfX_i, \bfX_j)$, $\widetilde{\bfK}_{\bff_i \bff_i} = \bfK_{\bff_i \bff_i} + \sigma_{\epsilon}^2 \bfI_{n_i}$. 
$\bfK_{\calA (\cdot) } = 
[ 
    \cov[\mu_i(\cdot), y_*(\cdot)] 
]_{i=1}^{M}$  
$=$ 
$\bfK_{ (\cdot) \calA }^{\top} \in \bbR^{M}$, $\bfK_{\calA \calA}= \big[ \cov[\mu_i(\cdot), \mu_j(\cdot)] \big]_{i,j=1}^{M} \in \bbR^{M \times M}$, $\bfmu_{\calA} = \big[ \mu_1(\cdot), \ldots, \mu_M(\cdot) \big]^{\top} \in \bbR^M$.

Among these methods, PoE, gPoE, BCM, and rBCM are computationally efficient but assume independence between experts, leading to inconsistent predictions as the dataset size grows. The grBCM method addresses consistency by introducing a global expert but requires training augmented models $\mathcal{M}_i^+$ for $i=2,\ldots,M$, each on datasets of size $n_c + n_i$, significantly increasing computational cost. NPAE accounts for correlations through the full covariance matrix $\mathbf{K}_{\calA \calA}$ in \cref{eq:npae}, achieving consistency but at $O(M^3)$ cost for inverting this matrix, which becomes prohibitive for large $M$.

These limitations motivate our approach: we seek an aggregation method that (1) incorporates expert correlations for consistency, (2) maintains computational efficiency comparable to simple baselines, and (3) naturally applies to both exact GP and SVGP settings. Our key insight introduced in \Cref{sec:DistGP with optimal weights} is to leverage the optimized combination technique (OptiCom) from sparse grid theory, which is introduced in \Cref{sec:GP with OptiCom}, to derive optimal aggregation weights that address these requirements.

\section{GP regression with OptiCom}
\label{sec:GP with OptiCom}
This section introduces GP regression using the optimized combination technique (OptiCom). We first review the standard combination technique (CT) in \Cref{subsec:CT}, followed by OptiCom in \Cref{subsec:OptiCom}, and then show how it is applied to GP regression.

\subsection{Combination technique (CT)}
\label{subsec:CT}
\textit{Combination technique} (CT) \citep{griebel1990combination,hegland2007combination}, first introduced in \citep{smolyak1963quadrature}, is an efficient tool for approximating the sparse grid spaces. When partial projection operators commute, as in interpolation with tensor product spaces, the combination technique provides the exact sparse grid solution. Sparse grids of level $\eta$ and dimension $d$ are defined as \citep{garcke2013sparse}
\begin{equation}\label{eq:sgd}
    \Omega_{\eta}^{d} 
    := \bigcup_{\vert \underl \vert_1 = \eta+d-1} \Omega_{\underl} 
    = \bigcup_{\vert \underl \vert_1 = \eta+d-1} \Omega_{l_1}^1 \times \cdots \times \Omega_{l_d}^1.
\end{equation}
Here, $\underl = (l_1,\ldots,l_d)$, $\vert \underl \vert_1 = \sum_{j=1}^d l_j$, where $l_j \in \bbN^+$ for all $j=1,\ldots d$. $\Omega_{l_1}^1 \times\cdots\times \Omega_{l_d}^1 = \{ (\omega_{l_1},\ldots,\omega_{l_d}) \vert \omega_{l_j} \in \Omega_{l_j}^1 \; \text{for} \; j \in \{1,\ldots,d\} \}$ denotes the n-ary Cartesian product over $d$ one-dimensional set $\Omega_{l_j}^1 \subset \bbR$, $j=1,\ldots,d$. We suppose $\Omega_{l_j}^1$ consists of $2^{l_j-1}$ uniformly distributed points over a fixed interval, i.e., $\Omega_{l_j}^1$ over the interval $[0,1]$ is given by $\Omega_{l_j}^1 = \left\{ \frac{i}{2^{l_j}}: i=1,\ldots,2^{l_j-1} \right\}$.

Each set of grids $\Omega_{\underl} \subset \bbR^d$ is associated with a piecewise $d$-linear basis function $\bfphi_{\underl, \underh }(\cdot)$ defined as:
\begin{equation}
    \bfphi_{\underl,\underh}(\bfx) := \prod_{j=1}^d \phi_{l_j, h_j}(x_j), \quad h_j = 1,\ldots, 2^{l_j-1},
    \quad 
    \phi_{l_j,h_j}(x) = \begin{cases}
        1-\lvert 2^{l_j} x - h_j \rvert, & x \in \big[ \frac{h_j-1}{2^{l_j}}, \frac{h_j+1}{2^{l_j}} \big], \\
        0, & \text{otherwise},
    \end{cases}
\end{equation}
where $\bfx = (x_1,\ldots,x_d) \in \bbR^d$ is a $d$-dimensional point, $\underh=(h_1,\ldots,h_d)$, $\phi_{l_j,h_j}(\cdot)$ is a piecewise linear hierarchical basis shown in \Cref{fig:one-dim basis for sparse grid space}.

These basis functions can define function spaces $V_{\underl} := \text{span}\{ \bfphi_{\underl,\underh}: h_j=1,\ldots,2^{l_j}-1, j=1,\ldots,d \}$ on the grids $\Omega_{\underl}$, then the hierarchical increment spaces $W_{\underl}$ can be defined by
\begin{gather}
    W_{\underl} := \text{span}\{ \bfphi_{\underl,\underh}: \underh \in\B_{\underl} \}, \\
    \B_{\underl} := \{ \underh \in \bbN^{d} :h_j=1,\ldots,2^{l_j}-1, h_j \text{ is odd for all } j=1,\ldots,d \}.
\end{gather}
Therefore, the function spaces can be represented by the hierarchical increment spaces $V_{\underl} = \bigoplus_{\underk\; \leq \; \underl} W_{\underk}$ and each function $f_{\underl} \in V_{\underl}$ can be uniquely represented by $f_{\underl}(\bfx) = \sum_{\vert\underl\vert_1 \leq \eta+d-1} \sum_{\underh \in \B_{\underl}} \alpha_{\underl,\underh} \bfphi_{\underl,\underh}(\bfx)$ with hierarchical coefficients $\alpha_{\underl,\underh} \in \bbR$. The space $\oplus_{i=1}^{n} S_i$ is the direct sum of the spaces $S_1,\ldots, S_n$ if and only if all nonzero vectors $s_i \in S_i$ (for $i = 1,\ldots,n$) are linearly independent, and $\underk \leq \underl$ indicates that each component of $\underk$ is less than or equal to the corresponding component of $\underl$. The combination technique linearly combines the discrete solutions $f_{\underl}(\bfx)$ from the partial grids $\Omega_{\underl}$ according to the combination formula (see \Cref{fig:ct}):
\begin{equation}\label{eq:ct}
    f_{\eta}^{c}(\bfx) = \sum_{\eta \leq \vert \underl \vert_1 \leq \eta+d-1} (-1)^{\eta+d-1-\vert\underl\vert_1} \binom{d-1}{\vert\underl\vert_1 - \eta} f_{\underl}(\bfx),
\end{equation}
where the function $f_{\eta}^{c}$ is in the sparse grid space $V_{\eta}^{s} := \bigoplus_{\vert\underl\vert_1 \leq \eta+d-1} W_{\underl}$.

\subsection{Optimized combination technique (OptiCom)}
\label{subsec:OptiCom}
\textit{Optimized Combination Technique} (OptiCom)
\citep{hegland2002additive,garcke2006regression} selects the ``best possible'' combination coefﬁcients adaptively so that an optimal combination approximation is obtained. More specifically, we aim to minimize the functional
\begin{equation}\label{eq:opticom minimizer}
    J(c_1,\ldots,c_{b}) = \Vert \calP_{\eta}^{s} f - \sum_{i=1}^{b} c_i \calP_i f \Vert^2,
\end{equation}
where $\calP_{\eta}^{s} f$ denotes the projection into the sparse grid space $V_{\eta}^{s}$, $\calP_i f$ denotes the projection into one of the spaces $V_{\underl}$ in \cref{eq:ct}, $b$ is the number of summed terms in combination technique formula \cref{eq:ct}. Using simple expansions and formula $\langle\calP_{\eta}^{s} f, \calP_i f\rangle = \langle\calP_i f, \calP_i f\rangle$, we can obtain
\begin{equation}\label{eq:opticom minimizer expansion}
    J(c_1,\ldots,c_{b}) = \sum_{i,j=1}^{b} c_i c_j \langle\calP_i f, \calP_j f\rangle - 2\sum_{i=1}^{b} c_i \Vert \calP_i f \Vert^2 + \Vert \calP_{\eta}^{s} f \Vert^2.
\end{equation}
By differentiating with respect to the combination coefficients $c_i$ and setting each of these derivatives to zero, we can derive the following linear system
\begin{equation}
    \begin{bmatrix}
        \Vert \calP_1 f \Vert^2 & \cdots & \langle\calP_{1} f, \calP_{b} f\rangle \\
        \langle\calP_{2} f, \calP_{1} f\rangle  & \cdots & \langle\calP_{2} f, \calP_{b} f\rangle \\
        \vdots & \ddots & \vdots\\
        \langle\calP_{b} f, \calP_{1} f\rangle & \cdots & \Vert \calP_{b} f \Vert^2
    \end{bmatrix}
    \begin{bmatrix}
        c_1\\
        c_2\\
        \vdots\\
        c_{b}
    \end{bmatrix} = 
    \begin{bmatrix}
        \Vert \calP_{1} f \Vert^2 \\
        \Vert \calP_{2} f \Vert^2\\
        \vdots\\
        \Vert \calP_{b} f \Vert^2
    \end{bmatrix}.
\end{equation}
Using the optimal coefficients $c_i$, the OptiCom formula for the sparse grid of level $\eta$ is then given by
\begin{equation}\label{eq:opticom}
    f_{\eta}^{c}(\bfx) = \sum_{\eta \leq \vert \underl \vert_1 \leq \eta+d-1} c_{\underl} f_{\underl}(\bfx).  
\end{equation}
For the regression problem \citep{garcke2006regression}, we are seeking the solution of the optimization problem $f^* = \arg\min_{f \in V} R(f)$ with
\begin{equation}\label{eq:min regularization}
      R(f) := 
      \sum_{i=1}^n ( f(\bfx_i) - y_i )^2 + \lambda \Vert \calS f\Vert_2^2,
\end{equation}
where $\calS$ is a linear operator, $\lambda$ is the tuning parameter that penalizes the flexibility of the model, $f^* \in V$ is the unknown function we aim to recover from the given dataset $\{(\bfx_i,y_i)\}_{i=1}^n$, $\bfx_i\in\bbR^d$, $y_i\in\bbR$. The scalar product in the setting is then defined as
\begin{equation}\label{eq:opticom RLS inner product}
    \langle\calP_{\underl} f, \calP_{\underk} f\rangle_{\text{RLS}} = 
    \sum_{i=1}^{n} f_{\underl}(\bfx_i) f_{\underk}(\bfx_i) + \lambda \langle \calS f_{\underl}, \calS f_{\underk} \rangle_2,
\end{equation}
so that the minimization \cref{eq:min regularization} is an orthogonal projection of $f^*$ into the space $V$, i.e. if $\Vert f -f^* \Vert_{\text{RLS}}^2 \leq \Vert g -f^* \Vert_{\text{RLS}}^2$ then $R(f)\leq R(g)$.

\subsection{GP with OptiCom}
\label{subsec:GP with OptiCom}
\subsubsection{Posterior distribution}
It's easy to extend OptiCom to GP regression. Suppose the inducing inputs $\bfZ=\{\bfz_i\}_{i=1}^{m}$ ($\bfz_i\in\bbR^d$) are sparse grids $\Omega_{\eta}^d$ of level $\eta$ and dimension $d$ defined in \cref{eq:sgd}, then the predictive mean and covariance in \cref{eq:svgp-optimal-conditional-gp} can be approximated in the form
\begin{subequations}\label{eq:opticom-svgp-conditional-gp}
    \begin{align}\label{eq:opticom-svgp-conditional-mean}
    \hat{\mu}_{\eta}^{c}(\cdot)= 
    \sum_{\eta \leq \vert \underl \vert_1 \leq \eta+d-1} c_{\underl} \,\cdot\, \sigma_{\epsilon}^{-2}
    \bfK_{(\cdot) \bfu_{\underl}} \bfM_{\bfu_{\underl}\bfu_{\underl}}^{-1} \bfK_{\bfu_{\underl} \bff} \bfy,
    \end{align}
    \begin{align}\label{eq:opticom-svgp-conditional-variance}
        \hat{k}_{\eta}^{c}(\cdot, \cdot') 
        = 
        \sum_{\eta \leq \vert \underl \vert_1 \leq \eta+d-1} (-1)^{\eta+d-1-\vert\underl\vert_1} \binom{d-1}{\vert\underl\vert_1 - \eta}
        \Big[
        k(\cdot, \cdot') - 
        \bfK_{(\cdot)\bfu_{\underl}} (\bfK_{\bfu_{\underl} \bfu_{\underl}}^{-1} - \bfM_{ \bfu_{\underl} \bfu_{\underl} }^{-1}) \bfK_{\bfu_{\underl} (\cdot')}
        \Big],
    \end{align}
\end{subequations}
where $\bfu_{\underl}$ is the vector of inducing variables corresponding to the inducing inputs $\Omega_{\underl}$ defined in \cref{eq:sgd}, $\bfK_{ (\cdot) \bfu_{\underl} }=k(\cdot,\Omega_{\underl}) = \bfK_{ \bfu_{\underl} (\cdot) }^{\top}$, $\bfK_{\bfu_{\underl} \bfu_{\underl}}=k(\Omega_{\underl},\Omega_{\underl})$, and $\bfM_{\bfu_{\underl} \bfu_{\underl}} = \big[ \bfK_{\bfu_{\underl} \bfu_{\underl}} + \sigma_{\epsilon}^{-2} \bfK_{\bfu_{\underl} \bff} \bfK_{\bff \bfu_{\underl}} \big]$. To ensure that the predictive covariance remains consistent with that of the full GP, we use the CT coefficients rather than those from OptiCom in the covariance computation. 

\subsubsection{Optimal coefficients}
To compute the optimal coefficients $c_{\underl}$ in GP regression, we need to compute the inner product with respect to the  \textit{reproducing kernel Hilbert space} (RKHS) \citep{aronszajn1950theory}. Let $\calH$ be a RKHS endowed with an inner product $\langle \cdot,\cdot \rangle_{\calH}$, then the inner product between functions $f(\cdot)=\sum_{i=1}^{n}\alpha_i k(\cdot,\bfx_i)$ and $g(\cdot)=\sum_{j=1}^{n'}\alpha'_j k(\cdot,\bfx'_j)$ is given by $\langle f, g \rangle_{\calH} = \sum_{i=1}^{n} \sum_{j=1}^{n'} \alpha_i \alpha'_j k(\bfx_i,\bfx_j')$. In GP regression, the optimization problem in \cref{eq:min regularization} becomes $R(f) = \sum_{i=1}^n ( f(\bfx_i) - y_i )^2 + \lambda \Vert f\Vert_{\calH}^2$, where $\lambda=\sigma_{\epsilon}^2$ is the noise variance. Here the projection into one of the spaces $V_{\underl}$ has the form
\begin{equation}
    \calP_{\underl} f = \sigma_{\epsilon}^{-2} \bfK_{(\cdot) \bfu_{\underl}} \bfM_{\bfu_{\underl} \bfu_{\underl}}^{-1} \bfK_{\bfu_{\underl} \bff} \bfy = \sigma_{\epsilon}^{-2} k(\cdot, \Omega_{\underl}) \bfalpha_{\underl} = \sigma_{\epsilon}^{-2} \sum_{i\in\mathbf{I}_{\underl}} k(\cdot, \bfz_{i}) \alpha_i := f_{\underl}(\cdot),
\end{equation}
where $\bfI_{\underl} \subset \{1,\ldots,m\}$ is the set of indices of the inducing inputs $\{\bfz_i\}_{i=1}^{m}$ such that $\Omega_{\underl} = \{\bfz_i: i\in \bfI_{\underl}\}$, $\alpha_i$ is the $i$-th component of the vector $\bfalpha_{\underl} = \bfM_{ \bfu_{\underl} \bfu_{\underl} }^{-1} \bfK_{\bfu_{\underl} \bff} \bfy$. The scalar product defined in \cref{eq:opticom RLS inner product} becomes
\begin{align}
    \langle\calP_{\underl} f, \calP_{\underk} f\rangle_{\text{RLS}} 
    & = \sum_{i=1}^{n} f_{\underl}(\bfx_i) f_{\underk}(\bfx_i) + \lambda \langle f_{\underl}, f_{\underk} \rangle_{\calH} \nonumber\\
    & = \sigma_{\epsilon}^{-4} \sum_{i=1}^{n} k(\bfx_i, \Omega_{\underl}) \bfalpha_{\underl}  k(\bfx_i, \Omega_{\underk}) \bfalpha_{\underk} + \lambda \sigma_{\epsilon}^{-4}  \; \bfalpha_{\underl}^{\top} k(\Omega_{\underl}, \Omega_{\underk}) \bfalpha_{\underk} \nonumber\\
    & = \sigma_{\epsilon}^{-2} \bfalpha_{\underl}^{\top} \big[ \sigma_{\epsilon}^{-2} \bfK_{\bfu_{\underl} \bff} \bfK_{\bff \bfu_{\underk}} + \bfK_{\bfu_{\underl} \bfu_{\underk}} \big] \bfalpha_{\underk} := \sigma_{\epsilon}^{-2} \bfalpha_{\underl}^{\top} \bfM_{ \bfu_{\underl} \bfu_{\underk} } \bfalpha_{\underk},
\end{align}
where $\bfM_{ \bfu_{\underl} \bfu_{\underk} } = \big[ \bfK_{\bfu_{\underl} \bfu_{\underk}} + \sigma_{\epsilon}^{-2} \bfK_{\bfu_{\underl} \bff} \bfK_{\bff \bfu_{\underk}} \big]$. The details are outlined in \Cref{alg:gp opticom coeff} and \Cref{alg:gp opticom posterior}.

\begin{figure}[ht]
\centering
\begin{subfigure}[b]{0.46\textwidth}
    \includegraphics[width=\linewidth]{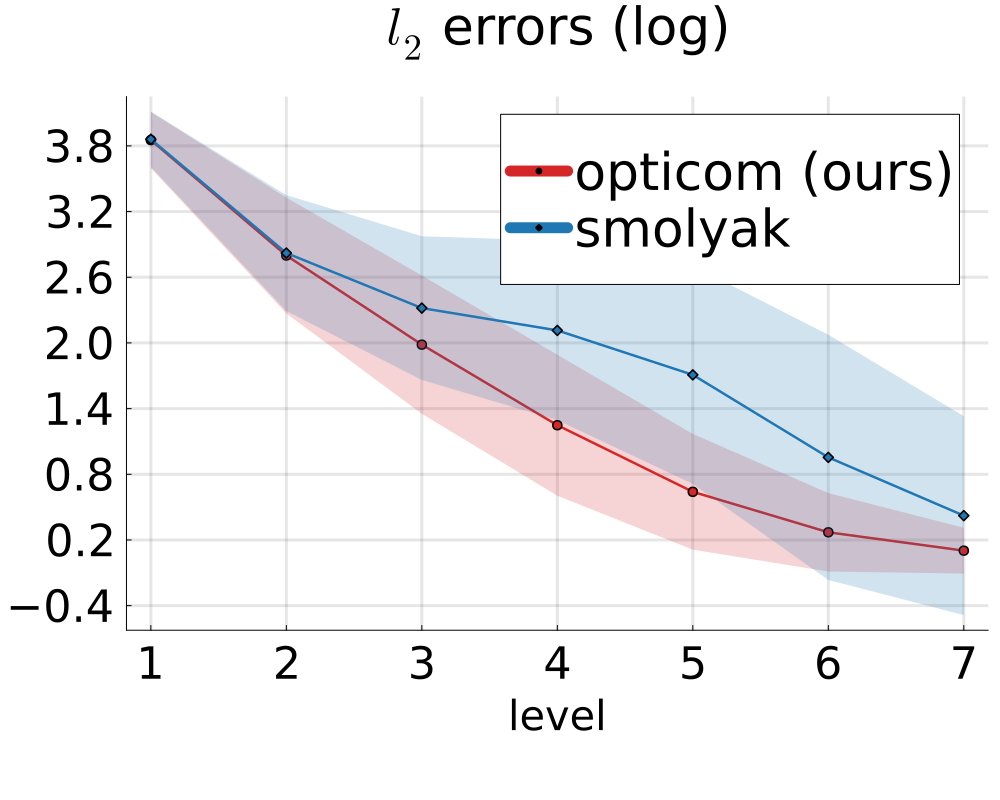}
\end{subfigure}\hspace{2em} 
\begin{subfigure}[b]{0.46\textwidth}
    \includegraphics[width=\linewidth]{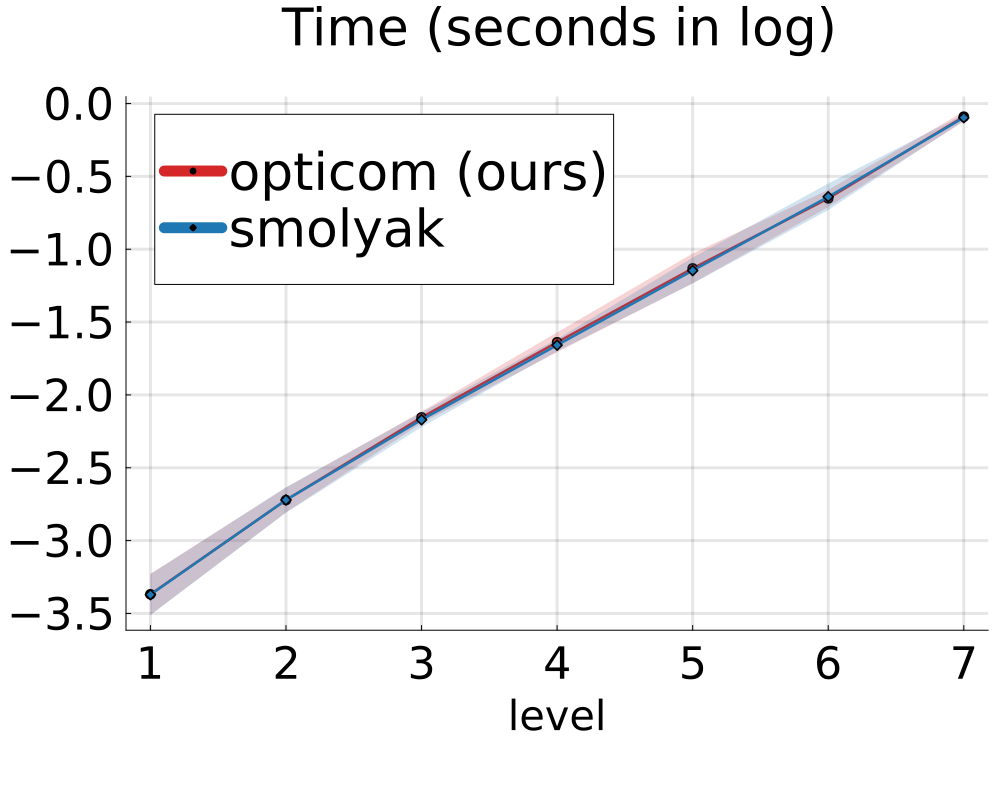}
\end{subfigure}
\caption{Errors and time for posterior mean of zero mean and Mat\'ern 3/2 GP with OptiCom and CT of dimension $d=2$ and level $\eta=1,2,\ldots,7$ tested on the Griewank function \citep{griewank1981generalized} over $n=2000$ training points and $n_t=1000$ test points. We denote GP with OptiCom by circles and GP with CT by diamonds. \textit{Left:} Logarithm of $l_2$ errors between GP posteriors and ground truth values averaged over 10 different lengthscales $[0.2, 0.4, 0.6, 0.8, 1.0, 1.2, 1.4, 1.6, 1.8, 2.0]$. \textit{Right:} Logarithm of time taken to compute the posterior mean averaged over $1000$ replications.}
\label{fig:gp opticom 2-dim}
\end{figure}

\subsubsection{Complexity}
From Lemma 6 in \citep{garcke2013sparse}, we know that the number of summed terms in combination technique formula \cref{eq:ct} is $b=\calO(\eta^{d-1})$, the size of each summed term is of order $m_b=\calO(2^{(\eta+d-1)})$, and dimension of the sparse grid space $V_{\eta}^s$ defined in \cref{eq:ct}, i.e., the number of grid points, is $\vert V_{\eta}^s \vert = \calO(2^{\eta} \cdot \eta^{d-1}) = \calO\big( h_{\eta}^{-1} \cdot \log(h_{\eta}^{-1})^{d-1} \big)$, where $h_{\eta}=2^{-\eta}$ is the mesh size of the sparse grids $\Omega_{\eta}^d$. The additional operations compared to GP with CT include computing $\bfM_{ \bfu_{\underl} \bfu_{\underk} }$ when $\underl \neq \underk$ and solving a $b \times b$ linear system to obtain the optimal coefficients, which costs $\calO(b^2 m_b^2 n)$ and $\calO(b^3)$ time respectively. Note that computation of $\{ \bfM_{ \bfu_{\underl} \bfu_{\underk} } \}_{\underl, \underk \in \bbI(\eta,d)}$ and $\{ \text{chol}(\bfM_{ \bfu_{\underl} \bfu_{\underl} }) \}_{\underl \in \bbI(\eta,d)}$ can be parallelized and reduced to $\calO(m_b^2 n)$ time and $\calO(m_b^3)$ time respectively, where $\bbI(\eta,d)=\{\underl: \eta \leq \vert \underl \vert_1 \leq \eta+d-1 \}$. Thus the computational complexity of the parallelized \Cref{alg:gp opticom posterior} is $\calO(m_b^3 + b^3 + m_b^2 n)=\calO(b^3+m_b^2 n)$. Although the computational complexity of the parallelized GP with OptiCom is larger than that of parallelized GP with CT theoretically, which is $\calO(m_b^2 n)$, $b$ is very small in most cases, hence $\calO(b^3+m_b^2 n) \approx \calO(m_b^2 n)$. \Cref{fig:gp opticom 2-dim} also demonstrates that GP with OptiCom achieves lower errors and smaller variance  than GP with CT at nearly the same time cost.

\section{Distributed GP with optimal weights}
\label{sec:DistGP with optimal weights}
In our previous formulation of GP with OptiCom, the inducing points were placed on fixed sparse grids rather than being learned through variational inference. While this design simplifies computation, it can limit performance when the sparse grid resolution is low, as the fixed locations may not adequately capture complex input structures. To address this limitation, we extend OptiCom to distributed GPs, which allows for greater flexibility and improved accuracy while maintaining computational efficiency and does not rely on sparse grid structures. Following the assumptions \Cref{subsubsec:Distributed GP training}, we consider the global training dataset $\calD = \{\bfX, \bfy\}$ of size $n$, partitioned into $M$ local datasets $\calD_i = \{ \bfX_i, \bfy_i \}$ of size $n_i$ for $i=1,\ldots,M$.  All the local datasets are assumed to share the same hyperparameters $\bftheta$.

\subsection{Distributed SVGP}
\subsubsection{Optimal weights}
To obtain the optimal weights, as shown the \cref{eq:opticom minimizer} and \cref{eq:opticom minimizer expansion} in \Cref{subsec:OptiCom}, the objective is to minimize
\begin{equation}\label{eq:min obj svgp}
    J(\beta_1,\ldots,\beta_M) = \Vert f - \sum_{i=1}^M \beta_i \calP_i f \Vert,
\end{equation}
where $f(\cdot) \sim\calGP(0, k(\cdot,\cdot))$ as defined in \Cref{subsubsec:Distributed GP training}, $\calP_if = f \vert \bfy_i:= f_i$ is the posterior conditioned on the local dataset $\calD_i=\{\bfX_i, \bfy_i\}$, and its distribution takes the form of \cref{eq:svgp-optimal-conditional-gp}. Similar to \Cref{subsec:GP with OptiCom}, the optimal weights $\{\beta_i\}_{i=1}^{M}$ can be obtained by solving the following linear system:
\begin{equation}\label{eq:linear system distribute svgp}
    \begin{bmatrix}
        \Vert \calP_1 f \Vert^2 & \cdots & \langle\calP_{1} f, \calP_{M} f\rangle \\
        \langle\calP_{2} f, \calP_{1} f\rangle  & \cdots & \langle\calP_{2} f, \calP_{M} f\rangle \\
        \vdots & \ddots & \vdots\\
        \langle\calP_{M} f, \calP_{1} f\rangle & \cdots & \Vert \calP_{M} f \Vert^2
    \end{bmatrix}
    \begin{bmatrix}
        \beta_1\\
        \beta_2\\
        \vdots\\
        \beta_{M}
    \end{bmatrix} = 
    \begin{bmatrix}
        \Vert \calP_{1} f \Vert^2 \\
        \Vert \calP_{2} f \Vert^2\\
        \vdots\\
        \Vert \calP_{M} f \Vert^2
    \end{bmatrix}.
\end{equation}
To reduce computational complexity, we randomly select a central dataset $\calD_c:= \{\bfX_c, \bfy_c\} \subset \calD=\{\bfX,\bfy\}$ of size $n_c \ll n$. In practice, we select one point from each local dataset $\calD_i$ and combine them to form the central dataset, resulting in $n_c=M$. Consequently, the scalar product is approximated as follows:
\begin{align}
    \langle\calP_{l} f, \calP_{k} f\rangle_{\text{RLS}} 
    & = \sum_{\bfx_i \in \bfX} f_{l}(\bfx_i) f_{k}(\bfx_i) + \lambda \langle f_{l}, f_{k} \rangle_{\calH} \nonumber\\
    & \approx \sum_{\bfx_i \in \bfX_c} f_{l}(\bfx_i) f_{k}(\bfx_i) + \lambda \langle f_{l}, f_{k} \rangle_{\calH} \nonumber\\
    & = \sigma_{\epsilon}^{-2} \bfalpha_{l}^{\top} \big[ \sigma_{\epsilon}^{-2} \bfK_{\bfu \bff_c} \bfK_{\bff_c \bfu} + \bfK_{\bfu \bfu} \big] \bfalpha_{k} := \sigma_{\epsilon}^{-2} \bfalpha_{l}^{\top} \bfM_{ \bfu \bfu }^{(c)} \bfalpha_{k},
\end{align}
where $\bfalpha_{l} = [ \bfM_{ \bfu \bfu }^{(l)} ]^{-1} \bfK_{\bfu \bff_l} \bfy_l$, $\bfM_{ \bfu \bfu }^{(l)} = \big[ \bfK_{\bfu \bfu} + \sigma_{\epsilon}^{-2} \bfK_{\bfu \bff_l} \bfK_{\bff_l \bfu} \big]$, $\bfK_{\bfu \bfu} = k(\bfZ, \bfZ)$, $\bfK_{\bfu \bff_l}=k(\bfZ, \bfX_l) = \bfK_{\bff_l \bfu}^{\top}$, $\bfM_{ \bfu \bfu }^{(c)} = \big[ \bfK_{\bfu \bfu} + \sigma_{\epsilon}^{-2} \bfK_{\bfu \bff_c} \bfK_{\bff_c \bfu} \big]$, $\bfK_{\bfu \bff_c}=k(\bfZ, \bfX_c) = \bfK_{\bff_c \bfu}^{\top}$, $\bfZ= \{\bfz_i\}_{i=1}^{m}$ are the inducing inputs corresponding to the optimized inducing variables $\bfu \in \bbR^m$. The details are outlined in \Cref{alg:distributed svgp optimal weights}.

In addition, the optimal weights computed in \cref{eq:linear system distribute svgp} inherently capture the correlations between local experts. This mechanism prevents redundant information from being overemphasized when experts are highly correlated, thus leading to a more balanced aggregation. As a result, the aggregated prediction becomes more consistent with that of a full GP. This integration of expert correlations is key to enhancing both the performance and stability of distributed GP.

\subsubsection{Aggregated prediction}
Instead of relying on PoE-based models, we directly use the optimal weights $\{\beta_i\}_{i=1}^{M}$ to predict the aggregated GP models after distributed training. Note that the correlations between the experts are already incorporated into the optimal weights. Therefore, we use the weighted local variances to approximate the aggregated variance, reducing computational complexity. The joint mean and joint variance are given as follows:
\begin{subequations}\label{eq:aggreg pred svgp}
    \begin{align}
        \mu_{\calA}(\cdot) 
        = \sum_{i=1}^{M} \beta_i  \bbE[f_i (\cdot)] 
        = \sum_{i=1}^{M} \beta_i \sigma_{\epsilon}^{-2} \bfK_{(\cdot) \bfu} \bfM_{\bfu \bfu}^{(i) -1} \bfK_{\bfu \bff_i} \bfy_i,
    \end{align}
    \begin{align}
        \sigma_{\calA}^{2}(\cdot) 
        = \sum_{i=1}^{M} \beta_i^2 \Var[ f_i (\cdot), f_i (\cdot)] = \sum_{i=1}^{M} \beta_i^2  \bfK_{(\cdot)\bfu} [\bfM_{\bfu \bfu}^{(i)}]^{-1} \bfK_{\bfu (\cdot)},
    \end{align}
\end{subequations}
The details are outlined in \Cref{alg:distributed svgp aggreg pred}.

\subsubsection{Complexity}
In \Cref{alg:distributed svgp optimal weights}, the computation of $\{ \bfM_{\bfu \bfu}^{(i)} \}_{i=1}^{M}$ and $\{ \text{chol}(\bfM_{\bfu \bfu}^{(i)}) \}_{i=1}^{M}$ requires $\calO(m^2 n_i)$ time and $\calO(m^3)$ time, respectively, after parallelization, and solving the $M \times M$ linear system costs $\calO(M^3)$ time. Therefore, the time complexity of the parallelized \Cref{alg:distributed svgp optimal weights} is $\calO(m^2 n_i + m^3 + M^3) = \calO(m^2 n_i + M^3)$. For the aggregated prediction, the computational complexity of the joint mean and joint variance is $\calO(M)$. Thus, the computational complexity of the parallelized \Cref{alg:distributed svgp aggreg pred} is $\calO(m^2 n_i +M^3 + M) = \calO(m^2 n_i +M^3)$, where $m$ is the size of shared inducing variables $\bfu$ across the experts $\{ \calM_i \}_{i=1}^{M}$, $n_i$ is the size of the local dataset $\calD_i$ for expert $\calM_i$, ($i=1,\ldots, M$), and $M$ is the number of experts. The comparison of the time complexity of the aggregation models and full SVGP is presented in the first row of \Cref{tab:distributed gp complexity}.

\begin{table}[!ht]
    \caption{Complexity of the prediction for aggregation models. $m$ is the number of inducing points, $n$ is the size of the entire training dataset, $n_i$ is the size of the local dataset $\calD_i$ corresponding to the expert $\calM_i$, $i=1,\ldots.\,M$, $n_c$ is the size of the dataset $\calD_c$ corresponding to the expert $\calM_c$ for grBCM. $M$ is the number of experts.}
    \label{tab:distributed gp complexity}
    \centering
    \resizebox{\textwidth}{!}{
    {\Large
    \begin{tabular}{c|c|cccccccc}
    \specialrule{2.5pt}{1pt}{1pt}
    \rowcolor{gray!20}
    \hline
         &  & PoE & gPoE & BCM & rBCM & grBCM & NPAE & \textbf{ opt (ours) } \\ 
        \specialrule{1.5pt}{1pt}{1pt}
         \multirow{2}{*}{Time} & SVGP & 
         $\calO(m^2 n_i+M)$ & 
         $\calO(m^2 n_i+M)$ & 
         $\calO(m^2 n_i+M)$ & 
         $\calO(m^2 n_i+M)$ & 
         $\calO(m^2 (n_i+n_c)+M)$ &
         $\calO(m^2 n_i M + M^3)$ & 
         $\calO(m^2 n_i + M^3)$\\
         ~ & ExactGP & 
         $\calO(n_i^3+M)$ & $\calO(n_i^3+M)$ & $\calO(n_i^3+M)$ & $\calO(n_i^3+M)$ & $\calO((n_i+n_c)^3 + M)$ &
         $\calO(n_i^3 M + M^3)$ & 
         $\calO(n_i^3 + M^3)$\\
        \specialrule{2.5pt}{1pt}{1pt}
    \end{tabular}
    }
    }
\end{table}

\subsection{Distributed exact GP}
The optimal weights method for distributed GPs is also applicable to exact GPs. In this case, the distribution of $\calP_i f = f \vert \bfy_i := f_i$ follows the form given in \cref{eq:conditional-gp}. Consequently, the scalar product is expressed as:
\begin{align}
    \langle\calP_{l} f, \calP_{k} f\rangle_{\text{RLS}} 
    & \approx \sum_{\bfx_i \in \bfX_c} f_{l}(\bfx_i) f_{k}(\bfx_i) + \lambda \langle f_{l}, f_{k} \rangle_{\calH} \nonumber\\
    & = \bfalpha_{l}^{\top} \big[ \bfK_{\bff_l \bff_c} \bfK_{\bff_c \bff_k} + \bfK_{\bff_l \bff_k} \big] \bfalpha_{k} :=  \bfalpha_{l}^{\top} \bfM_{ \bff_l \bff_k }^{(c)} \bfalpha_{k},
\end{align}
where $\bfalpha_{l} = \widetilde{\bfK}_{\bff_l \bff_l}^{-1} \bfy_l$, $\widetilde{\bfK}_{\bff_l \bff_l} = \bfK_{\bff_l \bff_l} + \sigma_{\epsilon}^2 \bfI_{n_l}$, $\bfM_{ \bff_l \bff_k }^{(c)} = \big[ \bfK_{\bff_l \bff_k} + \bfK_{\bff_l \bff_c} \bfK_{\bff_c \bff_k} \big]$, $\bfK_{\bff_l \bff_k}=k(\bfX_l, \bfX_k)$. Therefore, the joint mean and joint variance are given by:
\begin{align}\label{eq:aggreg pred exactgp}
    \mu_{\calA}(\cdot)
    = \sum_{i=1}^{M} \beta_i  \bfK_{(\cdot) \bff_i} \widetilde{\bfK}_{\bff_i \bff_i}^{-1} \bfy_i, 
    \quad
    \sigma_{\calA}^{2}(\cdot) 
    = \sum_{i=1}^{M} \beta_i^2 \bfK_{(\cdot)\bff_i} \widetilde{\bfK}_{\bff_i \bff_i}^{-1} \bfK_{\bff_i(\cdot)}.
\end{align}
The detailed procedures are outlined in \Cref{alg:distributed exactgp optimal weights} and \Cref{alg:distributed exactgp aggreg pred}. The time complexity of the parallelized \Cref{alg:distributed exactgp aggreg pred} is $\calO(n_i^3+M^3)$, where $n_i$ is the size of the local dataset $\calD_i$ for expert $\calM_i$, and $M$ is the number of experts. A comparison of the time complexity of the aggregation models and the full exact GP is presented in the last row of \Cref{tab:distributed gp complexity}.

\subsection{Convergence}
In this section, we introduce the theoretical properties of the proposed aggregation method with optimal weights. In contrast to PoE and BCM families whose weights are chosen heuristically, our weights are defined
as the minimizer of a function space residual objective in \cref{eq:min obj svgp} and are computed from the resulting linear system in \cref{eq:linear system distribute svgp}. This optimization-based method is what enables the consistency for the aggregated predictor.

\paragraph{Setting.}
Let the global dataset $\calD$ be randomly partitioned into local datasets $\{\calD_i\}_{i=1}^{M_n}$ without
replacement, where the number of experts $M_n$ increases with the global dataset size $n$ such that
(i) $\lim_{n \to \infty} M_n = \infty$, and (ii) $\liminf_{n \to \infty} \frac{n}{M_n^2} > 0$.
Then the local sample sizes satisfy $n_i \asymp n/M_n \to \infty$.
Assume each expert $\calM_i$ is a zero-mean GP with the same kernel and Gaussian noise model as the full GP.

For each GP expert, let $\calP_i f$ denote the posterior operator conditioned on the local dataset $\calD_i$
(as defined in \cref{eq:min obj svgp}). Let $\mu_i(\cdot)$ and $\sigma_i^2(\cdot)$ be the predictive mean and variance of expert $i$, where $i=1,\ldots,M_n$, and let $\mu_{\text{full}}(\cdot)$ and $\sigma_{\text{full}}^2(\cdot)$ denote the predictive mean and variance of the full GP trained on $\calD$.

\paragraph{Assumptions.}

\begin{enumerate}
    \item \textbf{local predictive consistency.}
    
    We assume a standard GP predictive consistency property: as the local sample size $n_i\to\infty$, the predictive mean and variance of the $i$-th expert converge to the same limiting predictive quantities as the full GP, at any fixed test input $\bfx_*$. This is a classical result in GP posterior consistency theory \citep{choi2005posterior}, and it is also the auxiliary fact from proof of Proposition~2 in \citep{liu2018generalized} to control the asymptotic behavior of each expert's predictive mean and variance. Specifically, for any fixed $\bfx_*$, 
    \begin{align}\label{eq:local_consistency_mean_var}
        \lim_{n\to\infty} \max_{1\le i\le M_n} 
        \big|\mu_i(\bfx_*)-\mu_{\text{full}}(\bfx_*)\big| 
         = 0,
        \qquad
        \lim_{n\to\infty} \max_{1\le i\le M_n}
        \big| 
        \sigma_i^2(\bfx_*) -\sigma_{\text{full}}^2(\bfx_*) 
        \big| 
        = 0.
    \end{align}

    \item \textbf{weight stability.}
    
    We assume the optimizer $\bfbeta_n = (\beta_{1,n}, \ldots, \beta_{M_n, n})^{\top}$ returned by the linear system in \cref{eq:linear system distribute svgp} is stable, i.e.:
    \begin{align}
        \|\bfbeta_n\|_{\ell_1} \le B \quad \text{for all sufficiently large $n$},
        \label{eq:l1_stability}
    \end{align} 
    for some constant $B$ independent of $n$. This assumption is standard for linear estimators: without controlling the magnitude of $\bfbeta_n$, the residual term $\sum_{i=1}^{M_n} \beta_{i,n}(f-\calP_i f)$ below may fail to vanish even if each $(f-\calP_i f)\to 0$. In practice, such stability can be enforced by adding a small jitter regularizer to the Gram matrix in \cref{eq:linear system distribute svgp}.
\end{enumerate}

\paragraph{Aggregation prediction.}
The aggregated predictor of the proposed method has the form (see \cref{eq:aggreg pred svgp,eq:aggreg pred exactgp})
\begin{align}\label{eq:agg_mean_var_forms}
    \mu_{\calA}(\bfx_*)=\sum_{i=1}^{M_n}\beta_{i,n}\,\mu_i(\bfx_*),
    \qquad
    \sigma_{\calA}^2(\bfx_*)=\sum_{i=1}^{M_n}\beta_{i,n}^2\,\sigma_i^2(\bfx_*).
\end{align}
Define the sum of weights $s_n:=\sum_{i=1}^{M_n}\beta_{i,n}$, then we have the decomposition as follows:
\begin{align}\label{eq:mean_decomp}
    \mu_{\calA}(\bfx_*) = s_n\,\mu_{\text{full}}(\bfx_*) + \sum_{i=1}^{M_n}\beta_{i,n}\big(\mu_i(\bfx_*)-\mu_{\text{full}}(\bfx_*)\big).
\end{align}
Thus, mean consistency reduces to proving (i) $s_n\to 1$, and (ii) the second term in \cref{eq:mean_decomp} converges to zero.

\paragraph{Mean.}
\begin{itemize}
    \item \textbf{Step 1: the objective value at the optimizer converges to zero.}

    Define the function space residual objective from \cref{eq:min obj svgp} 
    \begin{align}\label{eq:Jn_def}
        J_n(\bfbeta)\;:=\;\Big\| f-\sum_{i=1}^{M_n}\beta_i\,\calP_i f\Big\|.
    \end{align}
    Let $\bfbeta_n$ be a minimizer of $J_n(\bfbeta)$. Then for any $j\in\{1,\dots,M_n\}$,
    \begin{align}\label{eq:min_le_single_expert}
        J_n(\bfbeta_n)\;\le\;J_n(\bfe_j)\;=\;\|f-\calP_j f\|,
    \end{align}
    where $\bfe_j$ is the $j$-th standard basis vector. By Assumption~1 and $n_j\asymp n/M_n\to\infty$, we have 
    \begin{align}
        \lim_{n\to\infty} \|f-\calP_j f\| = 0
    \end{align}
    Therefore,
    \begin{align}\label{eq:Jn_to_0}
        \lim_{n\to\infty} J_n(\bfbeta_n) = 0
    \end{align}

    \item \textbf{Step 2: derive $s_n=\sum_{i=1}^{M_n} \beta_{i,n}\to 1$.}

    Using the identity
    \begin{align}\label{eq:key_identity}
        f-\sum_{i=1}^{M_n}\beta_{i,n}\calP_i f
        =(1-s_n)f+\sum_{i=1}^{M_n}\beta_{i,n}\big(f-\calP_i f\big),
    \end{align}
    we bound $|1-s_n|$ by applying the triangle inequality:
    \begin{align}\label{eq:sn_bound}
        |1-s_n| \cdot \|f\|
        &=\big\|(1-s_n)f\big\| \nonumber\\
        &\le \Big\|f-\sum_{i=1}^{M_n}\beta_{i,n}\calP_i f\Big\|
        +\Big\|\sum_{i=1}^{M_n}\beta_{i,n}(f-\calP_i f)\Big\| \nonumber\\
        &\le J_n(\bfbeta_n)+\sum_{i=1}^{M_n}|\beta_{i,n}| \cdot \|f-\calP_i f\| \nonumber\\
        &\le J_n(\bfbeta_n)+\|\bfbeta_n\|_{\ell_1}\,\max_{1\le i\le M_n}\|f-\calP_i f\|.
    \end{align}
    By \cref{eq:Jn_to_0}, the first term $J_n(\bfbeta_n)$ tends to $0$. By Assumption~1, $\lim_{n\to\infty}\max_i\|f-\calP_i f\| = 0$,
    and by Assumption~2, $\|\bfbeta_n\|_{\ell_1}\le B$. Hence the right hand side of \cref{eq:sn_bound} converges to $0$,
    which implies that
    \begin{align}\label{eq:sum_beta_to_1}
        \lim_{n\to\infty} \sum_{i=1}^{M_n}\beta_{i,n}
        = \lim_{n\to\infty} s_n = 1.
    \end{align}

    \item \textbf{Step 3: prove mean consistency.}
    Return to \cref{eq:mean_decomp}. The second term is bounded by
    \begin{align}
        \Big|\sum_{i=1}^{M_n}\beta_{i,n}\big(\mu_i(\bfx_*)-\mu_{\text{full}}(\bfx_*)\big)\Big|
        \le \|\bfbeta_n\|_{\ell_1}\,\max_{1\le i\le M_n}\big|\mu_i(\bfx_*)-\mu_{\text{full}}(\bfx_*)\big|
        \xrightarrow[]{}0
        \quad
        \text{as}
        \quad
        n\to\infty
    \end{align}
    using \cref{eq:l1_stability,eq:local_consistency_mean_var}. Combining this with \cref{eq:sum_beta_to_1} yields
    \begin{align}
        \lim_{n\to\infty} 
        \mu_{\calA}(\bfx_*) 
        &=  \lim_{n\to\infty} \Big( s_n\,\mu_{\text{full}}(\bfx_*) + \sum_{i=1}^{M_n}\beta_{i,n}\big(\mu_i(\bfx_*)-\mu_{\text{full}}(\bfx_*)\big) \Big)
        \\
        & = 1 \cdot \mu_{\text{full}}(\bfx_*) + 0
        \\
        & = \mu_{\text{full}}(\bfx_*).
    \end{align}
    Therefore, the aggregated predictive mean is consistent.
\end{itemize}

\paragraph{Variance.}
Using \cref{eq:agg_mean_var_forms} we get the following decomposition for aggregated predictive variance:
\begin{align}\label{eq:var_decomp}
    \sigma_{\calA}^2(\bfx_*)
    &= \sum_{i=1}^{M_n}\beta_{i,n}^2 \sigma_i^2(\bfx_*) \nonumber\\
    &= \Big(\sum_{i=1}^{M_n}\beta_{i,n}^2\Big)\sigma_{\text{full}}^2(\bfx_*)
    + \sum_{i=1}^{M_n}\beta_{i,n}^2\big(\sigma_i^2(\bfx_*)-\sigma_{\text{full}}^2(\bfx_*)\big).
\end{align}
If $\|\bfbeta_n\|_{\ell_2}$ is uniformly bounded, then the second term in \cref{eq:var_decomp} goes to zero by Assumption~1. However, to obtain $\sigma_{\calA}^2(\bfx_*)\to\sigma_{\text{full}}^2(\bfx_*)$ one further needs the stronger condition $\sum_i\beta_{i,n}^2\to 1$, which we do not enforce. Consequently, our theoretical guarantee is primarily for the predictive mean, while variance consistency would require additional constraints on $\bfbeta_n$.

\paragraph{Stability.}
Assumption~2 is required for a rigorous argument, i.e.,  the term $\sum_i \beta_{i,n}(f-\calP_i f)$ in \cref{eq:key_identity} may not vanish without controlling the magnitude of $\bfbeta_n$ even when each component $(f-\calP_i f)\to 0$. In our method, $\bfbeta_n$ is obtained by solving the finite dimensional linear system \cref{eq:linear system distribute svgp} derived from the first-order optimality conditions of the function space objective $J_n(\bfbeta)$. The solution is stable whenever the corresponding Gram matrix is well-conditioned. In practice, we ensure numerical stability by adding a small
jitter to \cref{eq:linear system distribute svgp}, which is standard in GP computations and typically yields bounded solutions.

\section{Experiments}
\label{sec:exp}
In this section, we evaluate the proposed algorithm against full GP and several baselines that aggregate predictions via weighted averaging: PoE, gPoE, BCM, rBCM, and grBCM, using both synthetic data (\Cref{subsec:syn}) and UCI datasets \citep{asuncion2007uci} (\Cref{subsec:uci}). We restrict baselines to aggregation methods because our contribution is an aggregation algorithm for distributed GP experts, scalable methods for centralized GP such as inducing points approximations, structured kernel interpolation, Vecchia framework, or stochastic dual descent solvers are complementary and can be used within each expert, but they do not provide a competing aggregation mechanism under the distributed learning setting. The size of the inducing variables is set to $m=128$. We use the \textit{radial basis function} (RBF) kernel with separate lengthscales $l_i$ for each input dimension $i$, and $\sigma_f^2 > 0$ is the signal variance of the kernel. 
The code for the experiments is available at \url{https://github.com/hchen19/optimal-distgp}.

\subsection{Synthetic data}
\label{subsec:syn}
\subsubsection{Training}
\begin{figure}[hp!]
    \centering
    \begin{minipage}{\textwidth}
        \centering
        \includegraphics
        [height=0.115\textheight]{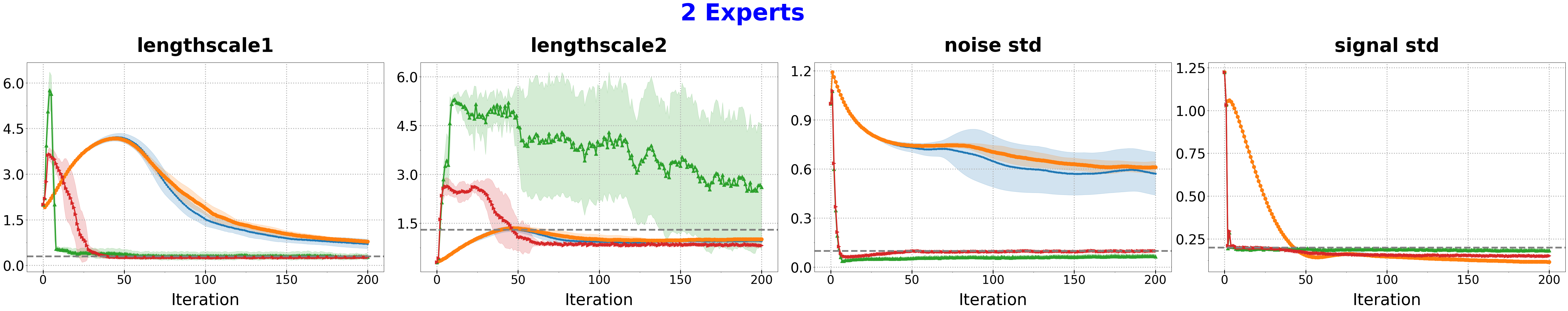}
    \end{minipage}
    \vfill
    \begin{minipage}{\textwidth}
        \centering
        \includegraphics
        [height=0.115\textheight]{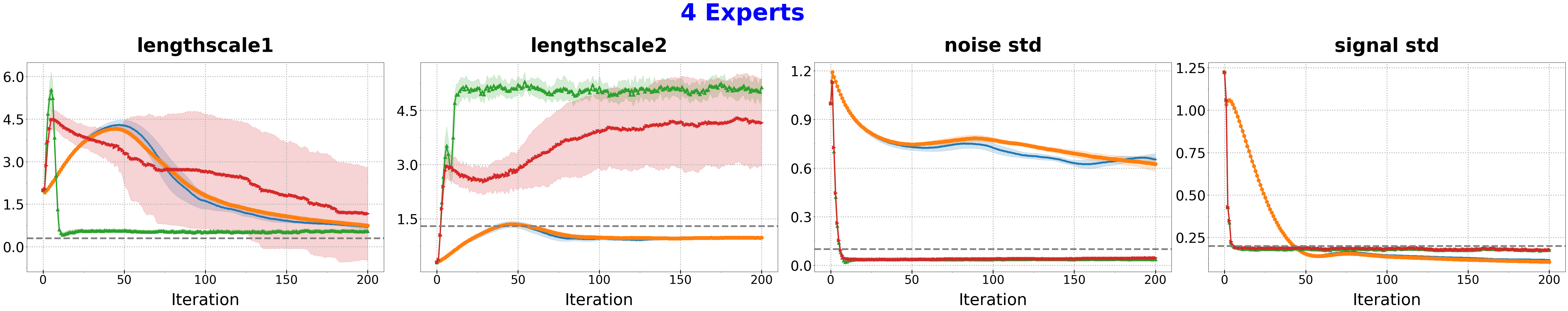}
    \end{minipage}
    \vfill
    \begin{minipage}{\textwidth}
        \centering
        \includegraphics
        [height=0.115\textheight]{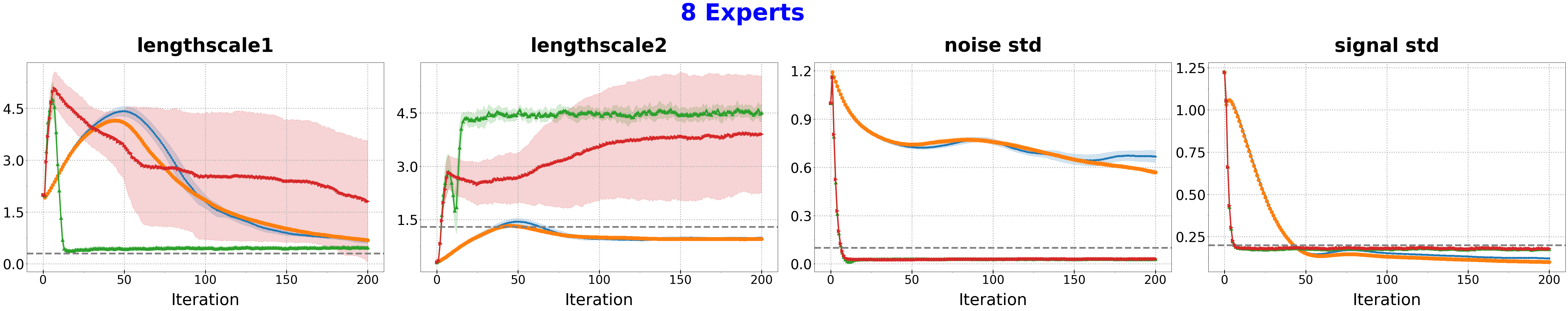}
    \end{minipage}
    \vfill
    \begin{minipage}{\textwidth}
        \centering
        \includegraphics
        [height=0.115\textheight]{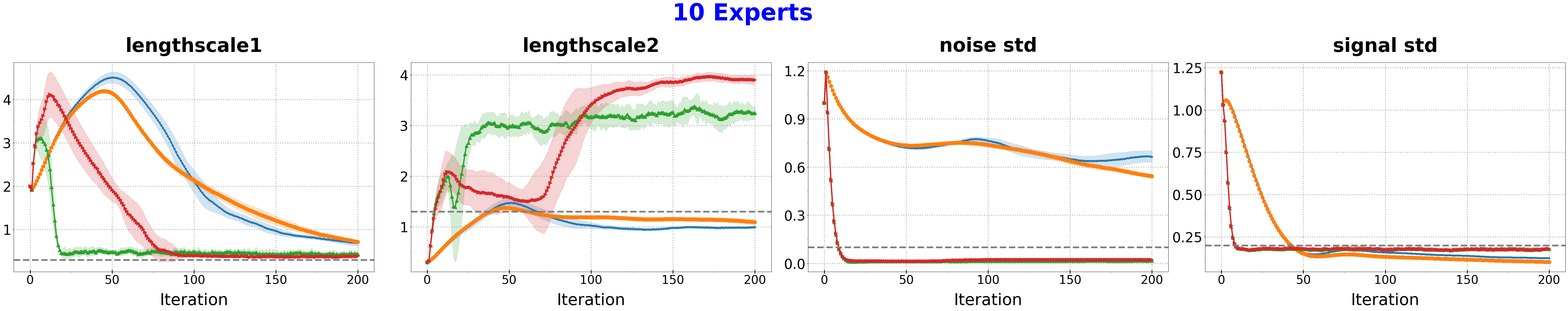}
    \end{minipage}
    \vfill
    \begin{minipage}{\textwidth}
        \centering
        \includegraphics
        [height=0.115\textheight]{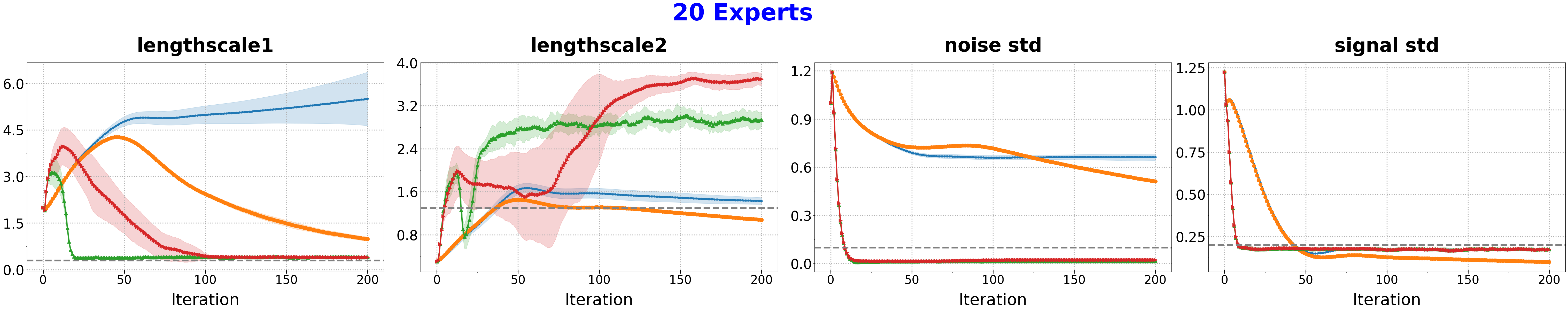}
    \end{minipage}
    \vfill
    \begin{minipage}{\textwidth}
        \centering
        \includegraphics
        [height=0.115\textheight]{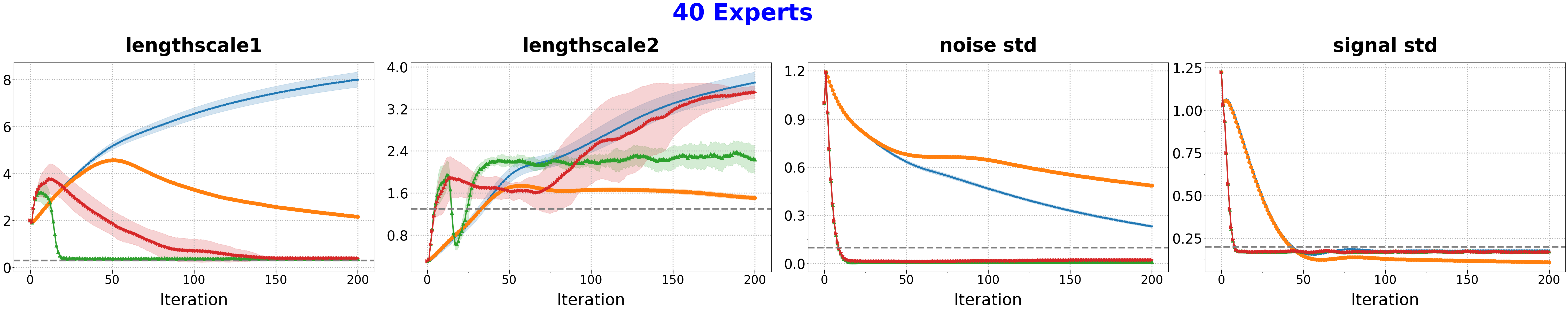}
    \end{minipage}
    \vfill
    \begin{minipage}{\textwidth}
        \centering
        \includegraphics
        [height=0.115\textheight]{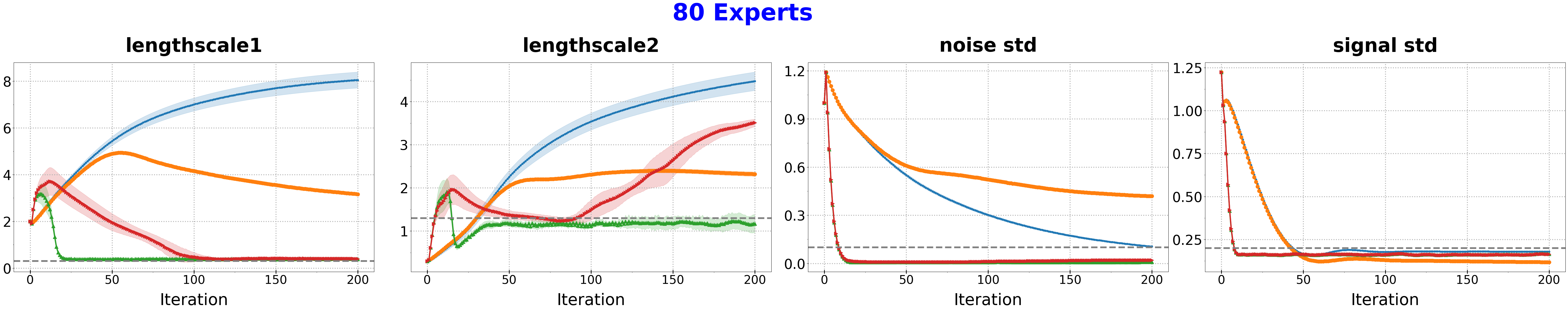}
    \end{minipage}
    \vfill
    \begin{minipage}{\textwidth}
        \centering
        \includegraphics
        [height=0.135\textheight]{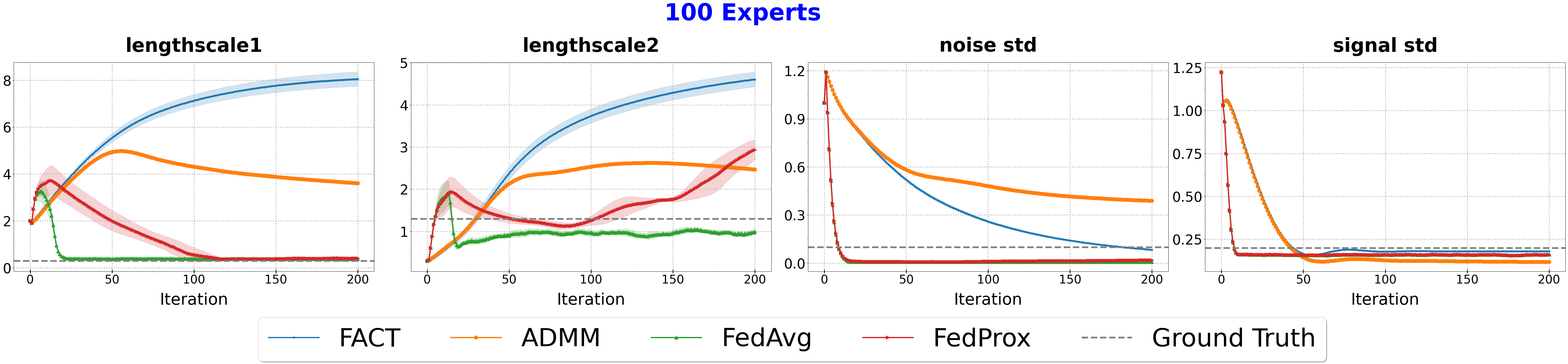}
    \end{minipage}
    \caption{Hyperparameter estimates versus the  training iterations on the $n=10^4$ dataset.}
    \label{fig:train_hyperparam_iter}
\end{figure}

In this section, we assess the performance of four distributed training methods for SVGP: FACT, ADMM, FedAvg, and FedProx. We generate a training dataset of size $n=10^4$ and dimension $d=2$ using the RBF kernel with $(l_1, l_2, \sigma_{\epsilon}, \sigma_f)=(0.3, 1.3, 0.1, 0.2)$. The dataset is distributed among $M=2, 4, 8, 10, 20, 40, 80, 100$ experts, respectively. We set the initial values of the hyperparameters $(l_1^{(0)}, l_2^{(0)}, \sigma_{\epsilon}^{(0)}, \sigma_f^{(0)})=(2.0, 0.3, 1.0, \sqrt{1.5})$, and use Adam optimizer \citep{kingma2014adam} with a learning rate of $0.1$ for all methods. For ADMM, the regularization parameter is set to $\rho=500$. For FedProx, the proximal coefficient is set to $\mu=0.01$, with a sampling rate of 0.5 for the local updates. The hyperparameters of the SVGP $\bftheta=(l_1, l_2, \sigma_{\epsilon}, \sigma_f, \bfu)^{\top}$ are trained over 200 iterations, where $\bfu$ are the inducing variables of size $m=128$. The results are averaged over 10 different seeds for each setting of $M$. 

\Cref{fig:train_hyperparam_iter} illustrates the change of the parameters with the training iterations for different numbers of experts $M$. We observe that FACT and ADMM perform worse as the number of experts $M$ increases, while FedAvg performs the best overall, converging faster and more stably than FedProx. \Cref{fig:train_hyperparam_boxplots} shows boxplots of the hyperparameter estimates after 200 iterations, indicating that FedAvg converges closest to the ground truth with the least variability. Therefore, we use FedAvg for distributed training in the experiments with the UCI datasets.

\begin{figure}[ht]
    \centering
\includegraphics[width=\linewidth]{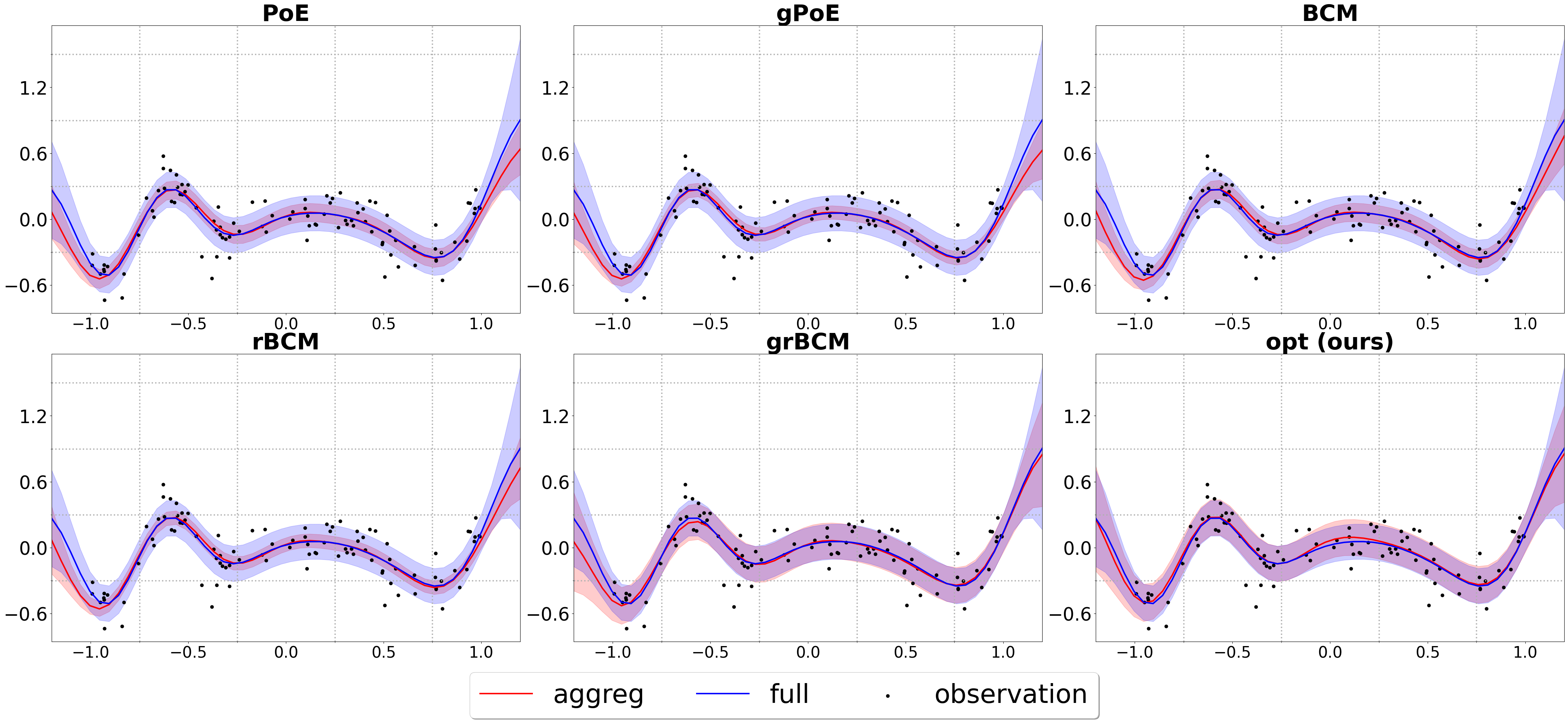}
\caption{SVGP posteriors with the different aggregation models on $n=100$ training points in $[-1,1]$ and $n_t=50$ test points in $[-1.2,1.2]$ with the number of experts $M=4$, dimension $d=1$, and the number of inducing variables $m=128$. We denote the aggregated predictions (aggreg) by dashed lines, the full SVGP posteriors (full) by solid lines,
and the observations by dots.}
\label{fig:distgp_full_aggreg_compare}
\end{figure}

\subsubsection{Prediction}
In this section, we evaluate the performance of aggregation models and full GP models without optimizing any hyperparameters. We generate the ground truth dataset using RBF kernel with fixed hyperparameters $(l_i, \sigma_{\epsilon}^2, \sigma_f^2) = (0.3, 0.25, 1.0)$ for all $i=1,\ldots,d$. For SVGP, we use $m=128$ inducing points, randomly generated with a fixed seed across all aggregation methods.

\Cref{fig:distgp_full_aggreg_compare} shows the aggregated predictions of SVGP on $n=100$ training points uniformly sampled in $[-1,1]$ and $n_t=50$ test points evenly spaced in $[-1.2, 1.2]$, with dimension $d=1$ and $M=4$ experts. We compare our method (opt) against five baselines (PoE, gPoE, BCM, rBCM, grBCM) and the full SVGP model using the same inducing points. The proposed method yields predictions that closely match the full SVGP, especially in the extrapolation region outside the training domain (i.e., in $[-1.2, -1] \cup [1, 1.2]$). In contrast, most baselines show degraded performance in these regions. Furthermore, the predictive variances of our method and grBCM align more closely with the full SVGP, suggesting better consistency of the predictions compared to other methods.

\begin{figure}[ht]
\centering
\begin{subfigure}{\columnwidth}
    \includegraphics[width=\linewidth]{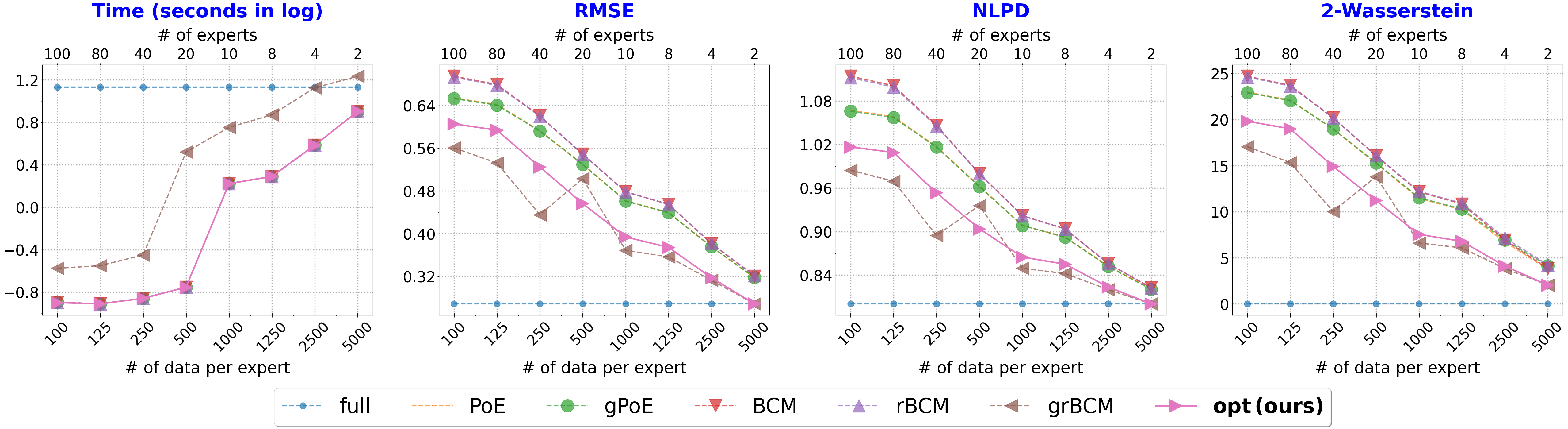}
    \caption{mean}
    \label{fig:distgp_syn_exactgp_plain}
\end{subfigure}
\begin{subfigure}{\columnwidth}
    \includegraphics[width=\linewidth]{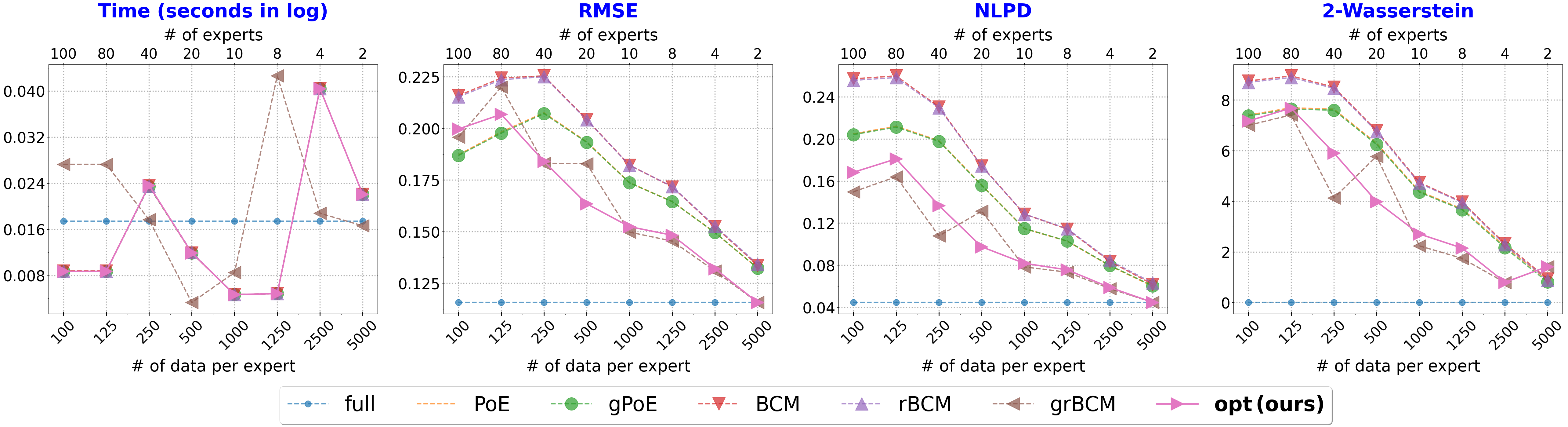}
    \caption{standard deviation}
    \label{fig:distgp_syn_exactgp_std}
\end{subfigure}
\caption{Comparison of the aggregation models for \textbf{Exact GP} with the RBF kernel in dimension $d=2$ on $n=10^4$ training points in $[-1,1]^2$ and $n_t=2500$ test points in $[-1.2,1.2]^2$. The number of experts considered are $M=2, 4, 8, 10, 20, 40, 80, 100$. The lower $x$-axis represents the size of the local training dataset $n_i = n / M$, and the upper $x$-axis represents the number of the experts $M$. \textit{Left:} Logarithm of time for computing the aggregated predictions. \textit{Middle Left:} RMSE between the aggregated predictions and the ground truth. \textit{Middle Right:} NLPD between the aggregated predictions and the ground truth. \textit{Right:} 2-Wasserstein distance to the full GP. \textit{Top:} Mean of the computational time, RMSE, NLPD, and 2-Wasserstein distance. \textit{Bottom:} Standard deviation of the computational time, RMSE, NLPD, and 2-Wasserstein distance. }
\label{fig:distgp_syn_exactgp}
\end{figure}

\begin{figure}[!ht]
\centering
\begin{subfigure}{\columnwidth}
    \includegraphics[width=\linewidth]{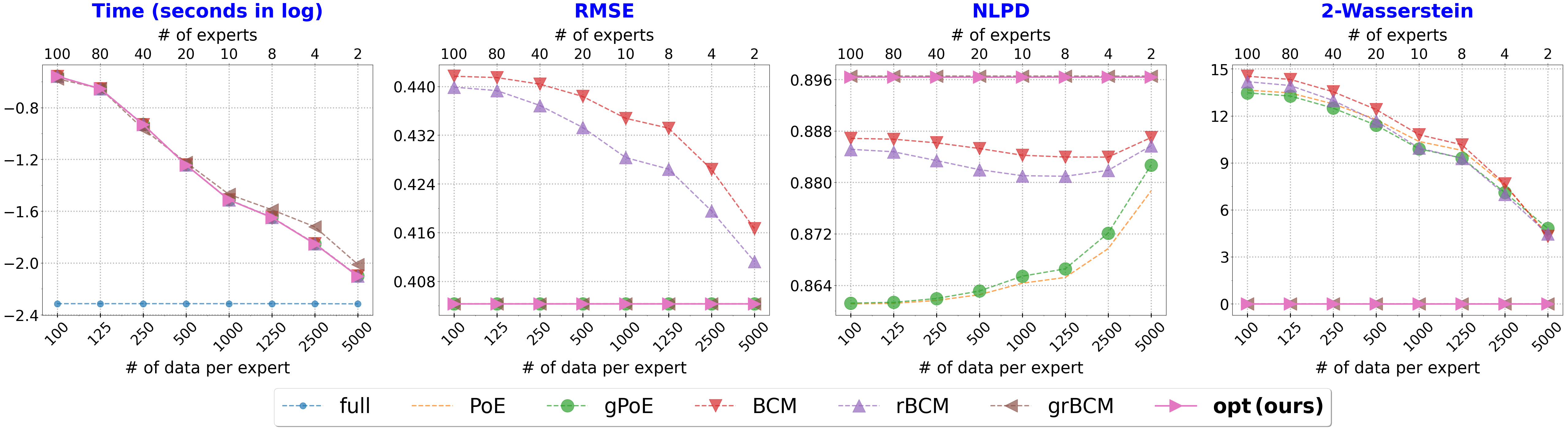}
    \caption{mean}
    \label{fig:distgp_syn_svgp_plain}
\end{subfigure}
\begin{subfigure}{\columnwidth}
    \includegraphics[width=\linewidth]{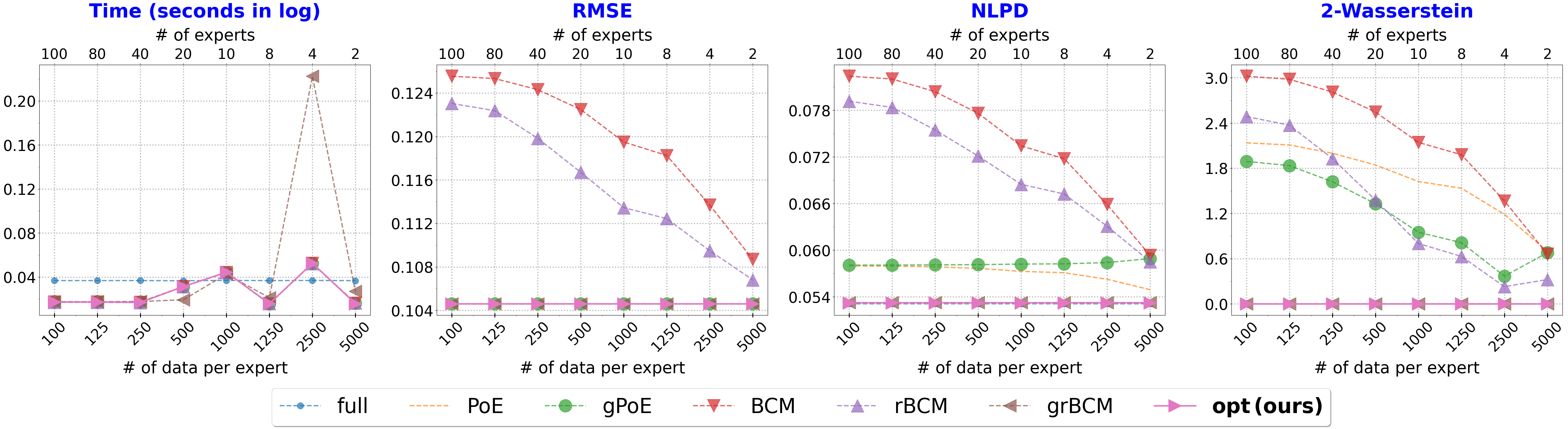}
    \caption{standard deviation}
    \label{fig:distgp_syn_svgp_std}
\end{subfigure}
\caption{Comparison of the aggregation models for \textbf{SVGP} with the RBF kernel in dimension $d=2$ on $n=10^4$ training points in $[-1,1]^2$, $m=128$ inducing points, and $n_t=2500$ test points in $[-1.2,1.2]^2$. The number of experts considered are $M=2, 4, 8, 10, 20, 40, 80, 100$. The lower $x$-axis represents the size of the local training dataset $n_i = n / M$, and the upper $x$-axis represents the number of the experts $M$. \textit{Left:} Logarithm of time for computing the aggregated predictions. \textit{Middle Left:} RMSE between the aggregated predictions and the ground truth. \textit{Middle Right:} NLPD between the aggregated predictions and the ground truth. \textit{Right:} 2-Wasserstein distance to the full GP. \textit{Top:} Mean of the computational time, RMSE, NLPD, and 2-Wasserstein distance. \textit{Bottom:} Standard deviation of the computational time, RMSE, NLPD, and 2-Wasserstein distance.}
\label{fig:distgp_syn_svgp}
\end{figure}

Next, we evaluate the aggregation models on $n=10^4$ training data of dimension $d=2$ in $[-1,1]^2$ and $n_t = 50 \times 50 = 2500$ test points, evenly spaced over the interval $[-1.2,1.2]^2$. The training dataset is divided among $M=2, 4, 8, 10, 20, 40, 80, 100$ experts, respectively. To eliminate the impact of seeds, we average all the metrics over 10 different seeds for each setting of $M$. 

\Cref{fig:distgp_syn_exactgp} compares the performance of various aggregation models for exact GPs using four metrics: computational time, \textit{root mean squared error} (RMSE), \textit{negative log predictive density} (NLPD), and 2-Wasserstein distance to the full exact GP (measuring how closely each aggregated model approximates the full GP). While grBCM performs the best in RMSE, NLPD, and 2-Wasserstein distance, it requires more computational time due to the larger size of the augmented local training datasets and has larger standard deviations of the metrics. Our method, GP with optimal weights, provides comparable predictions with significantly lower computational cost. \Cref{fig:distgp_syn_svgp} also compares four metrics for SVGP across all aggregation models. It's observed that our method and grBCM significantly outperform other methods for SVGP.

\subsection{UCI regression for temporal extrapolation}
\label{subsec:uci}
In this section, we evaluate the aggregation models on three real-world time-series datasets from the UCI repository \citep{asuncion2007uci}: Gas Turbine CO and NOx Emission \citep{kaya2019predicting}, Beijing PM2.5 \citep{liang2015assessing}, and Metro Interstate Traffic Volume \citep{hogue2019metro}. Unlike standard random train-test splits, we employ \textit{temporal extrapolation splits} where models are trained on earlier years and tested on future years, creating a more realistic and challenging evaluation scenario that tests the models' ability to handle distribution shift and regime changes.

\paragraph{Dataset descriptions and splits:}
\begin{itemize}
    \item \textbf{Gas Turbine CO \& NOx} (36,733 samples, 10 features): Hourly sensor measurements from 2011--2015. Training on 2011--2013, testing on 2014--2015. This dataset contains two prediction targets: CO and NOx emissions, we run experiments independently for each target. The future years contain operational regimes not fully represented in training data.
    \item \textbf{Beijing PM2.5} (43,824 samples, 13 features): Hourly air quality measurements from 2010--2014. Training on 2010--2013, testing on 2014. The test period includes extreme pollution events such as haze episodes that challenge model calibration.
    \item \textbf{Metro Traffic} (48,204 samples, 8 features): Hourly traffic volume measurements from 2012--2018. Training on 2012--2016, testing on 2017--2018. Traffic patterns exhibit regime shifts from construction, policy changes, and seasonal variations.
\end{itemize}

\paragraph{Experimental setup:}
To create a challenging distributed setting where expert diversity and correlation become important, we partition each training dataset into $M=20, 40, 60$ heterogeneous experts by randomly shuffling and dividing the data. Specifically, we use \textbf{non-shared hyperparameters} across experts (i.e., each expert optimizes its own lengthscale, kernel variance, and noise variance independently), which creates the strong correlated local datasets that our method can handle. We employ SVGP with $m=128$ inducing points and RBF kernel. All aggregation models utilize FedAvg with the Adam optimizer, with 500 training iterations to ensure sufficient convergence for the extrapolation task. Results are averaged over 10 different random seeds.

\Cref{fig:uci_extrapolation_errorbar} presents the performance metrics of the aggregation models on these four datasets. It's observed that: (1) BCM and rBCM perform poorly, particularly for NOx and Beijing PM2.5, where it yields significantly degraded RMSE and NLPD due to overconfident but incorrect experts being equally weighted. (2) While grBCM shows competitive RMSE, it requires substantially more computation time due to the need to train additional models. (3) Our method, distributed SVGP with optimal weights (opt), consistently achieves the best or near-best RMSE and NLPD across all datasets while maintaining computational efficiency comparable to simpler baselines. Notably, on the challenging extrapolation tasks (NOx and Metro Traffic), our method's ability to account for expert correlation and reliability provides clear advantages in both prediction accuracy and uncertainty quantification. Detailed metric values are reported in \Cref{tab:uci_extrapolation}.

\begin{figure}[hp!]
    \centering
    \begin{minipage}{\textwidth}
        \centering
        \includegraphics[height=0.16\textheight]{figs/distgp_opt/uci/gasturbine_co_errorbar.png}
    \end{minipage}
    \vfill
    \begin{minipage}{\textwidth}
        \centering
        \includegraphics[height=0.16\textheight]{figs/distgp_opt/uci/gasturbine_nox_errorbar.png}
    \end{minipage}
    \vfill
    \begin{minipage}{\textwidth}
        \centering
        \includegraphics[height=0.16\textheight]{figs/distgp_opt/uci/beijingpm25_errorbar.png}
    \end{minipage}
    \vfill
    \begin{minipage}{\textwidth}
        \centering
        \includegraphics[height=0.185\textheight]{figs/distgp_opt/uci/metrotraffic_errorbar.png}
    \end{minipage}
    \caption{Comparison of aggregation models for SVGP on temporal extrapolation tasks. Models are trained on earlier years and tested on future years to evaluate performance under distribution shift.}
    \label{fig:uci_extrapolation_errorbar}
\end{figure}

\begin{table}[H]
\caption{Results of aggregation models for SVGP on UCI datasets for temporal extrapolation tasks. Models are trained on earlier years and tested on future years. \colorbox{golden}{Best} and \colorbox{red!30}{worst} performances are highlighted in yellow and red, respectively.}
\label{tab:uci_extrapolation}
    \centering
    \resizebox{\textwidth}{!}{
    \begin{tabular}{c|c|cccccccc}
    \specialrule{2.5pt}{1pt}{1pt}
    \rowcolor{gray!20}
        Data & $M$ & Metric & SVGP & PoE & gPoE & BCM & rBCM & grBCM & \textbf{ opt (ours) } \\ 
        \specialrule{1.5pt}{1pt}{1pt}
        \multirow{9}{*}{\rotatebox[]{90}{\textbf{Gas Turbine CO}}}
 & \multirow{3}{*}{20}
 & Time & \cellcolor{red!30} 35.405 $\pm$ 0.415 & 0.164 $\pm$ 0.002 & 0.164 $\pm$ 0.001 & \cellcolor{golden} 0.163 $\pm$ 0.001 & 0.164 $\pm$ 0.001 & 0.297 $\pm$ 0.001 & \cellcolor{golden} 0.163 $\pm$ 0.001 \\
 &  & RMSE & 1.642 $\pm$ 0.053 & \cellcolor{golden} 1.518 $\pm$ 0.005 & 1.518 $\pm$ 0.005 & \cellcolor{red!30} 2.828 $\pm$ 0.054 & 2.510 $\pm$ 0.045 & 1.615 $\pm$ 0.117 & 1.519 $\pm$ 0.005 \\
 &  & NLPD & 1.955 $\pm$ 0.047 & 2.310 $\pm$ 0.009 & 2.157 $\pm$ 0.007 & \cellcolor{red!30} 5.881 $\pm$ 0.195 & 4.350 $\pm$ 0.123 & 1.991 $\pm$ 0.199 & \cellcolor{golden} 1.830 $\pm$ 0.004 \\
         \cline{2-10}
 & \multirow{3}{*}{40}
 & Time & \cellcolor{red!30} 35.405 $\pm$ 0.415 & 0.100 $\pm$ 0.008 & 0.095 $\pm$ 0.005 & 0.094 $\pm$ 0.005 & 0.093 $\pm$ 0.005 & 0.157 $\pm$ 0.005 & \cellcolor{golden} 0.091 $\pm$ 0.005 \\
 &  & RMSE & 1.656 $\pm$ 0.000 & 1.526 $\pm$ 0.002 & \cellcolor{golden} 1.525 $\pm$ 0.002 & \cellcolor{red!30} 3.106 $\pm$ 0.020 & 2.939 $\pm$ 0.019 & 1.733 $\pm$ 0.214 & 1.527 $\pm$ 0.004 \\
 &  & NLPD & 1.973 $\pm$ 0.000 & 2.366 $\pm$ 0.005 & 2.273 $\pm$ 0.006 & \cellcolor{red!30} 7.269 $\pm$ 0.085 & 6.170 $\pm$ 0.071 & 2.244 $\pm$ 0.378 & \cellcolor{golden} 1.833 $\pm$ 0.004 \\
         \cline{2-10}
 & \multirow{3}{*}{60}
 & Time & \cellcolor{red!30} 35.405 $\pm$ 0.415 & 0.079 $\pm$ 0.007 & 0.074 $\pm$ 0.005 & 0.072 $\pm$ 0.002 & 0.072 $\pm$ 0.002 & 0.109 $\pm$ 0.001 & \cellcolor{golden} 0.069 $\pm$ 0.001 \\
 &  & RMSE & 1.656 $\pm$ 0.000 & 1.550 $\pm$ 0.004 & \cellcolor{golden} 1.549 $\pm$ 0.005 & \cellcolor{red!30} 3.187 $\pm$ 0.021 & 3.078 $\pm$ 0.020 & 1.789 $\pm$ 0.196 & 1.552 $\pm$ 0.005 \\
 &  & NLPD & 1.973 $\pm$ 0.000 & 2.432 $\pm$ 0.010 & 2.360 $\pm$ 0.010 & \cellcolor{red!30} 7.748 $\pm$ 0.094 & 6.935 $\pm$ 0.077 & 2.370 $\pm$ 0.261 & \cellcolor{golden} 1.849 $\pm$ 0.005 \\
        \specialrule{1.5pt}{1pt}{1pt}
        \multirow{9}{*}{\rotatebox[]{90}{\textbf{Gas Turbine NOx}}}
 & \multirow{3}{*}{20}
 & Time & \cellcolor{red!30} 36.239 $\pm$ 1.322 & \cellcolor{golden} 0.164 $\pm$ 0.003 & 0.165 $\pm$ 0.003 & \cellcolor{golden} 0.164 $\pm$ 0.003 & 0.165 $\pm$ 0.003 & 0.299 $\pm$ 0.008 & \cellcolor{golden} 0.164 $\pm$ 0.003 \\
 &  & RMSE & \cellcolor{golden} 1.065 $\pm$ 0.041 & 1.070 $\pm$ 0.012 & 1.070 $\pm$ 0.012 & \cellcolor{red!30} 1.964 $\pm$ 0.035 & 1.949 $\pm$ 0.032 & 1.095 $\pm$ 0.083 & 1.066 $\pm$ 0.012 \\
 &  & NLPD & 4.103 $\pm$ 0.446 & 7.900 $\pm$ 0.171 & 7.898 $\pm$ 0.176 & \cellcolor{red!30} 26.145 $\pm$ 0.892 & 25.703 $\pm$ 0.802 & 5.297 $\pm$ 1.285 & \cellcolor{golden} 4.032 $\pm$ 0.100 \\
         \cline{2-10}
 & \multirow{3}{*}{40}
 & Time & \cellcolor{red!30} 36.239 $\pm$ 1.322 & 0.104 $\pm$ 0.007 & 0.099 $\pm$ 0.006 & 0.095 $\pm$ 0.004 & 0.094 $\pm$ 0.004 & 0.162 $\pm$ 0.008 & \cellcolor{golden} 0.092 $\pm$ 0.004 \\
 &  & RMSE & 1.115 $\pm$ 0.000 & 1.063 $\pm$ 0.008 & 1.063 $\pm$ 0.008 & \cellcolor{red!30} 2.036 $\pm$ 0.017 & 2.021 $\pm$ 0.016 & 1.096 $\pm$ 0.057 & \cellcolor{golden} 1.062 $\pm$ 0.013 \\
 &  & NLPD & 4.332 $\pm$ 0.000 & 7.990 $\pm$ 0.126 & 7.987 $\pm$ 0.122 & \cellcolor{red!30} 28.948 $\pm$ 0.476 & 28.460 $\pm$ 0.443 & 5.350 $\pm$ 0.713 & \cellcolor{golden} 3.890 $\pm$ 0.101 \\
         \cline{2-10}
 & \multirow{3}{*}{60}
 & Time & \cellcolor{red!30} 36.240 $\pm$ 1.322 & 0.084 $\pm$ 0.003 & 0.076 $\pm$ 0.005 & 0.075 $\pm$ 0.004 & 0.075 $\pm$ 0.003 & 0.112 $\pm$ 0.003 & \cellcolor{golden} 0.072 $\pm$ 0.002 \\
 &  & RMSE & 1.115 $\pm$ 0.000 & 1.066 $\pm$ 0.007 & 1.066 $\pm$ 0.007 & \cellcolor{red!30} 2.085 $\pm$ 0.020 & 2.069 $\pm$ 0.020 & \cellcolor{golden} 1.043 $\pm$ 0.102 & 1.065 $\pm$ 0.007 \\
 &  & NLPD & 4.332 $\pm$ 0.000 & 8.104 $\pm$ 0.112 & 8.100 $\pm$ 0.111 & \cellcolor{red!30} 30.656 $\pm$ 0.594 & 30.121 $\pm$ 0.565 & 5.759 $\pm$ 1.683 & \cellcolor{golden} 3.860 $\pm$ 0.053 \\
        \specialrule{1.5pt}{1pt}{1pt}
        \multirow{9}{*}{\rotatebox[]{90}{\textbf{Beijing PM2.5}}}
 & \multirow{3}{*}{20}
 & Time & \cellcolor{red!30} 60.044 $\pm$ 1.202 & 0.242 $\pm$ 0.007 & 0.240 $\pm$ 0.007 & 0.240 $\pm$ 0.007 & 0.240 $\pm$ 0.007 & 0.422 $\pm$ 0.008 & \cellcolor{golden} 0.239 $\pm$ 0.007 \\
 &  & RMSE & 7.780 $\pm$ 0.116 & 7.776 $\pm$ 0.016 & 7.775 $\pm$ 0.015 & \cellcolor{red!30} 8.530 $\pm$ 0.048 & 8.363 $\pm$ 0.039 & \cellcolor{golden} 7.755 $\pm$ 0.094 & 7.775 $\pm$ 0.018 \\
 &  & NLPD & 8.640 $\pm$ 0.652 & 35.010 $\pm$ 0.188 & 30.668 $\pm$ 0.299 & \cellcolor{red!30} 38.362 $\pm$ 0.289 & 31.270 $\pm$ 0.280 & 9.147 $\pm$ 0.602 & \cellcolor{golden} 7.350 $\pm$ 0.100 \\
         \cline{2-10}
 & \multirow{3}{*}{40}
 & Time & \cellcolor{red!30} 60.044 $\pm$ 1.202 & 0.138 $\pm$ 0.008 & 0.127 $\pm$ 0.003 & 0.125 $\pm$ 0.002 & 0.125 $\pm$ 0.002 & 0.228 $\pm$ 0.002 & \cellcolor{golden} 0.124 $\pm$ 0.002 \\
 &  & RMSE & \cellcolor{golden} 7.655 $\pm$ 0.000 & 7.830 $\pm$ 0.013 & 7.829 $\pm$ 0.013 & \cellcolor{red!30} 8.823 $\pm$ 0.055 & 8.682 $\pm$ 0.054 & 7.786 $\pm$ 0.119 & 7.831 $\pm$ 0.010 \\
 &  & NLPD & 9.020 $\pm$ 0.000 & 39.237 $\pm$ 0.137 & 36.061 $\pm$ 0.179 & \cellcolor{red!30} 46.846 $\pm$ 0.554 & 40.320 $\pm$ 0.481 & 11.787 $\pm$ 2.199 & \cellcolor{golden} 7.232 $\pm$ 0.096 \\
         \cline{2-10}
 & \multirow{3}{*}{60}
 & Time & \cellcolor{red!30} 60.044 $\pm$ 1.202 & 0.123 $\pm$ 0.016 & 0.095 $\pm$ 0.003 & \cellcolor{golden} 0.094 $\pm$ 0.002 & 0.095 $\pm$ 0.003 & 0.159 $\pm$ 0.004 & \cellcolor{golden} 0.094 $\pm$ 0.002 \\
 &  & RMSE & \cellcolor{golden} 7.655 $\pm$ 0.000 & 7.871 $\pm$ 0.012 & 7.869 $\pm$ 0.012 & \cellcolor{red!30} 9.041 $\pm$ 0.055 & 8.917 $\pm$ 0.047 & 7.942 $\pm$ 0.157 & 7.873 $\pm$ 0.013 \\
 &  & NLPD & 9.020 $\pm$ 0.000 & 41.091 $\pm$ 0.128 & 38.502 $\pm$ 0.151 & \cellcolor{red!30} 51.722 $\pm$ 0.562 & 45.666 $\pm$ 0.351 & 12.596 $\pm$ 1.516 & \cellcolor{golden} 7.140 $\pm$ 0.095 \\
        \specialrule{1.5pt}{1pt}{1pt}
        \multirow{9}{*}{\rotatebox[]{90}{\textbf{Metro Traffic}}}
 & \multirow{3}{*}{20}
 & Time & \cellcolor{red!30} 57.739 $\pm$ 2.806 & 0.317 $\pm$ 0.051 & 0.262 $\pm$ 0.010 & 0.255 $\pm$ 0.010 & 0.254 $\pm$ 0.009 & 0.467 $\pm$ 0.018 & \cellcolor{golden} 0.253 $\pm$ 0.010 \\
 &  & RMSE & 6.065 $\pm$ 0.701 & 5.612 $\pm$ 0.099 & \cellcolor{golden} 5.534 $\pm$ 0.060 & \cellcolor{red!30} 58.979 $\pm$ 2.695 & 34.593 $\pm$ 0.416 & 6.512 $\pm$ 1.272 & 5.741 $\pm$ 0.199 \\
 &  & NLPD & 7.874 $\pm$ 0.083 & 11.481 $\pm$ 0.157 & 7.960 $\pm$ 0.023 & \cellcolor{red!30} 235.124 $\pm$ 9.921 & 30.309 $\pm$ 1.069 & 7.942 $\pm$ 0.230 & \cellcolor{golden} 7.841 $\pm$ 0.026 \\
         \cline{2-10}
 & \multirow{3}{*}{40}
 & Time & \cellcolor{red!30} 57.739 $\pm$ 2.806 & 0.419 $\pm$ 0.118 & 0.162 $\pm$ 0.002 & 0.151 $\pm$ 0.004 & 0.146 $\pm$ 0.001 & 0.236 $\pm$ 0.003 & \cellcolor{golden} 0.144 $\pm$ 0.001 \\
 &  & RMSE & 6.236 $\pm$ 0.000 & 5.587 $\pm$ 0.027 & \cellcolor{golden} 5.569 $\pm$ 0.020 & \cellcolor{red!30} 49.724 $\pm$ 0.474 & 41.100 $\pm$ 0.278 & 6.050 $\pm$ 0.663 & 5.648 $\pm$ 0.067 \\
 &  & NLPD & 7.893 $\pm$ 0.001 & 16.096 $\pm$ 0.074 & 9.088 $\pm$ 0.020 & \cellcolor{red!30} 354.159 $\pm$ 3.765 & 76.711 $\pm$ 0.896 & 7.872 $\pm$ 0.145 & \cellcolor{golden} 7.843 $\pm$ 0.012 \\
         \cline{2-10}
 & \multirow{3}{*}{60}
 & Time & \cellcolor{red!30} 57.740 $\pm$ 2.806 & 0.330 $\pm$ 0.128 & 0.137 $\pm$ 0.004 & 0.122 $\pm$ 0.002 & 0.122 $\pm$ 0.002 & 0.167 $\pm$ 0.006 & \cellcolor{golden} 0.118 $\pm$ 0.001 \\
 &  & RMSE & 6.236 $\pm$ 0.000 & 5.651 $\pm$ 0.023 & \cellcolor{golden} 5.638 $\pm$ 0.020 & \cellcolor{red!30} 44.833 $\pm$ 0.618 & 40.345 $\pm$ 0.422 & 5.884 $\pm$ 0.406 & 5.692 $\pm$ 0.069 \\
 &  & NLPD & 7.893 $\pm$ 0.001 & 20.677 $\pm$ 0.092 & 10.498 $\pm$ 0.038 & \cellcolor{red!30} 433.313 $\pm$ 6.898 & 113.021 $\pm$ 0.620 & \cellcolor{golden} 7.829 $\pm$ 0.143 & 7.856 $\pm$ 0.012 \\
        \specialrule{2.5pt}{1pt}{1pt}
    \end{tabular}
    }
\end{table}

\section{Discussions}
\label{sec:conc}
\paragraph{Conclusions}
In this work, we introduced an algorithm to enhance GP regressions with inducing points on sparse grids using the optimized combination technique (OptiCom). Building on this foundation, we proposed a novel aggregated prediction algorithm suitable for both distributed exact GP and SVGP. This method leverages predictions from local models and optimally combines them using weights derived from solving a linear system. Additionally, it inherently incorporates correlations between experts through the computation of optimal weights. The proposed methods have a computational complexity requiring $\calO(M^3)$ additional time compared to PoE family and BCM family for $M$ GP experts. However, this overhead is minimal in practice, as the number of experts $M$ is typically small compared to the size of the local datasets. Through extensive experiments and comparisons with state-of-the-art aggregation models, we demonstrated the competitiveness and stability of our proposed method. Our algorithm consistently provides stable and accurate predictions, offering a robust solution for distributed GP regression.

\paragraph{Practical considerations}
Beyond model design, practical deployment involves key trade-offs in choosing the number of experts and inducing points. The choice of the number of experts $M$ involves a trade-off between efficiency and accuracy. Increasing $M$ improves computational efficiency, as each expert handles a smaller subset of the data and can be trained in parallel. However, if $M$ becomes too large, each expert receives too little data, degrading the quality of the local GP approximations. This effect is clearly illustrated in \Cref{fig:distgp_syn_exactgp}.
In practice, we recommend selecting $M$ based on empirical validation, such as cross-validation. A practical set of initial candidates can be defined as $M \in \{ \left\lfloor \sqrt{n}/i \right\rfloor : i = 4, 5, 6 \}$, where $n$ is size of the global training data. 

Similarly, the number of inducing points affects both accuracy and computational cost. More inducing points improve local approximations, and consequently lead to more accurate aggregated predictions, they also increase computational cost. In practice, selecting 5--10\% of the local training data size as the number of inducing points provides a good balance.

\paragraph{Limitations}
While our method guarantees consistent predictive means in the limit, it does not ensure consistent predictive variances. Theoretically, variance consistency would require the aggregation weights to satisfy an additional $\ell_2$-norm condition, which we do not impose. As a result, variance estimates may be slightly less accurate compared to methods like grBCM, especially in uncertainty-sensitive applications.

\paragraph{Future directions}
A direction for future work is to extend our framework such that it also ensures variance consistency, potentially by modifying the weight optimization process or incorporating additional constraints. Furthermore, integrating adaptive selection of expert sizes and inducing points could further enhance the scalability and flexibility of the method in large-scale settings.

\if0\blind{
\section*{Acknowledgements}
The authors acknowledge the generous support from the NSF grants DMS-2312173 and CNS-2328395. The experiments of this research were conducted on an A100 GPU provided by Texas A\&M High Performance Research Computing.} \fi

\section*{Data availability statement}
The data that support the findings of this study are openly available in \citep{asuncion2007uci} at \url{https://archive.ics.uci.edu/datasets}.

\bibliographystyle{apalike}
\spacingset{1}
\bibliography{IISE-Trans}

\appendix
\section{GP with OptiCom}
In this section, we provide supporting figures for \textit{optimized combination technique} (OptiCom) and detailed algorithms for GP with OptiCom in \Cref{subsec:gp opticom figs}
and \Cref{subsec:gp opticom algs} respectively.

\subsection{Figures}
\label{subsec:gp opticom figs}
\begin{figure}[hp!]
\centering
\begin{subfigure}[b]{0.45\textwidth}
    \includegraphics[height=7cm, width=\linewidth]{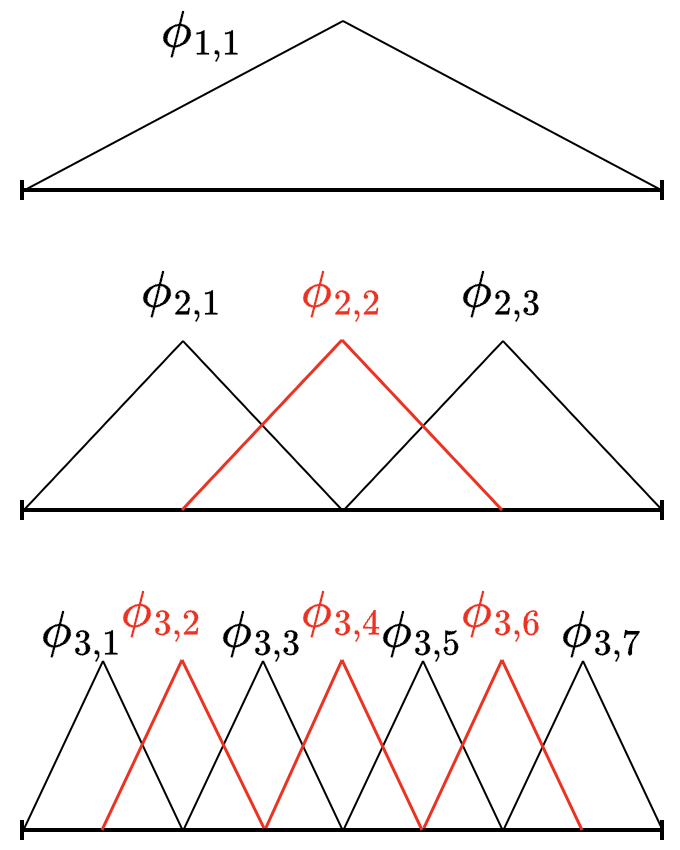}
    \caption[short]{Black lines: Hierarchical basis of level $\eta=3$ in one dimension.}
    \label{fig:one-dim basis for sparse grid space}
\end{subfigure}
\hspace{2em} 
\begin{subfigure}[b]{0.45\textwidth}
    \includegraphics[height=7cm, width=\linewidth]{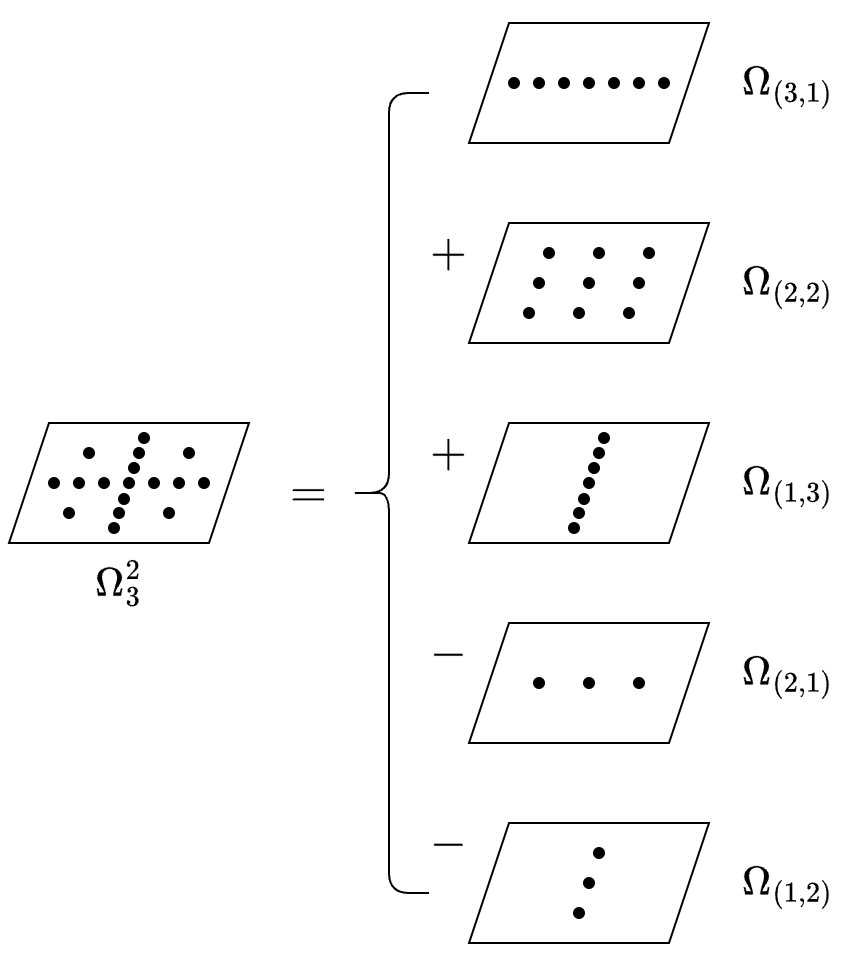}
    \caption[short]{Combination technique of level $\eta=3$ in $d=2$ dimensions: combine full grids $\Omega_{\underl}$, $\vert\underl\vert_1 \in \{3,4\}$, to get a sparse grid $\Omega_{\eta=3}^{d=2}$ corresponding to $V_3^{s}$.}
    \label{fig:two-dim sparse grid ct}
\end{subfigure}
\caption{Basis functions and combination technique of level $\eta=3$.}
\label{fig:ct}
\end{figure}

\subsection{Algorithms}
\label{subsec:gp opticom algs}
\begin{algorithm}[H]
\caption{Optimal Coefficients for GP with OptiCom}
\label{alg:gp opticom coeff}
\setstretch{0.99} 
\begin{algorithmic}[1]
    \STATE {\bfseries Input:} training inputs $\bfX=\{\bfx_i\}_{i=1}^n$, observations $\bfy=(y_1,\ldots,y_n)^{\top}$, noise variance $\sigma_{\epsilon}^{2}$, a GP denoted by $\calGP( \mu(\cdot), k(\cdot,\cdot'))$, sparse grids $\Omega_{\eta}^d$ of level $\eta$ and dimension $d$
    \STATE {\bfseries Output:} coefficients $\bfc = \{ c_{\underl^{(i)}} \}_{i=1}^{b}$, kernel weights $\{ \bfalpha_{\underl^{(i)}} \}_{i=1}^{b}$, Cholesky factors $\{ \bfL_{\underl^{(i)}} \}_{i=1}^{b}$, where $\{ \underl^{(i)} \}_{i=1}^{b} = \{\underl: \eta \leq \vert \underl \vert_1 \leq \eta+d-1 \}$
    \STATE $\bfA:=\text{zeros}(b,b)$ 
    \FOR[for loop over $t$ and $i$ can be parallelized]{$t=0$ {\bfseries to} $b-1$}
        \FOR[loop over the diagonal with offset $=t$]{$i=1$ {\bfseries to} $b-t$}
        \STATE $j := i+t$
        \STATE $\bfM_{} = \big[ k( \Omega_{\underl^{(i)}}, \Omega_{\underl^{(j)}} ) + \sigma_{\epsilon}^{-2} k( \Omega_{\underl^{(i)}}, \bfX) k( \bfX, \Omega_{\underl^{(j)}} ) \big]$
        \IF[$i=j$, $\bfM$ is symmetric and positive definite]{$t = 0$ }
            \STATE $\bfL_{\underl^{(i)}} := \text{chol}(\bfM)$ s.t. $\bfL_{\underl^{(i)}} \bfL_{\underl^{(i)}}^{\top} = \bfM$
            \STATE $\bfR :=\bfL_{\underl^{(i)}} \,\backslash\, \big[ k(\Omega_{\underl}, \bfX) \bfy \big]$
            \STATE $\bfalpha_{\underl^{(i)}} := \bfL_{\underl^{(i)}}^{\top} \,\backslash\, \bfR $
        \ENDIF
        \STATE $\bfA[i,j] = \bfA[j,i] = \sigma_{\epsilon}^{-2} \bfalpha_{\underl^{(i)}}^{\top} \bfM \bfalpha_{\underl^{(j)}}$
        \ENDFOR
    \ENDFOR
    \STATE $\bfc := \bfA \,\backslash\, \text{diag}(\bfA)$
    \STATE \textbf{return} $\bfc$, $\{ \bfalpha_{\underl^{(i)}} \}_{i=1}^{b}$, $\{ \bfL_{\underl^{(i)}} \}_{i=1}^{b}$
\end{algorithmic}
\end{algorithm}

\begin{algorithm}[H]
\caption{GP Posterior with OptiCom}
\label{alg:gp opticom posterior}
\setstretch{0.99} 
\begin{algorithmic}[1]
    \STATE {\bfseries Input:} training inputs $\bfX=\{\bfx_i\}_{i=1}^n$, observations $\bfy=(y_1,\ldots,y_n)^{\top}$, noise variance $\sigma_{\epsilon}^{2}$, test inputs $\bfX^* = \{ \bfx_i^* \}_{i=1}^{n_t}$, a GP denoted by $\calGP( \mu(\cdot), k(\cdot,\cdot'))$, sparse grids $\Omega_{\eta}^d$ of level $\eta$ and dimension $d$
    \STATE {\bfseries Output:} posterior mean $\hat{\mu}_{\eta}^{c}(\bfX^*)$ and posterior covariance $\hat{k}_{\eta}^{c}(\bfX^*, \bfX^*)$ in \cref{eq:opticom-svgp-conditional-gp}
    \STATE $\bfm := \text{zeros}(n_t)$; $\bfS := \text{zeros}(n_t, n_t)$
    \STATE get $\bfc = \{ c_{\underl^{(i)}} \}_{i=1}^{b}$, $\{ \bfalpha_{\underl^{(i)}} \}_{i=1}^{b}$, $\{ \bfL_{\underl^{(i)}} \}_{i=1}^{b}$ from \Cref{alg:gp opticom coeff}
    \FOR[for loop over $i$ can be parallelized]{$i=1$ {\bfseries to} $b$}
    \STATE $\bfm \leftarrow \bfm + c_{\underl^{(i)}} k(\bfX^*, \Omega_{\underl^{(i)}}) \bfalpha_{\underl^{(i)}}$
    \STATE $c_{\text{ct}}:= (-1)^{\eta+d-1-\vert\underl^{(i)}\vert_1} \binom{d-1}{\vert\underl^{(i)}\vert_1 - \eta}$
    \STATE $\bfP := k(\bfX^*, \Omega_{\underl^{(i)}}) \big[ k(\Omega_{\underl^{(i)}}, \Omega_{\underl^{(i)}}) \big]^{-1} k(\Omega_{\underl^{(i)}}, \bfX^*)$
    \STATE $\bfQ := \big[ k(\bfX^*, \Omega_{\underl^{(i)}}) \bfL_{\underl^{(i)}}^{-\top} \big] \big[ \bfL_{\underl^{(i)}}^{-1} k(\Omega_{\underl^{(i)}}, \bfX^*) \big]$
    \STATE $\bfS \leftarrow \bfS + c_{\text{ct}} \big[ k(\bfX^*, \bfX^*) - \bfP +\bfQ \big]$
    \ENDFOR
    \STATE $\hat{\mu}_{\eta}^{c}(\bfX^*) := \sigma_{\epsilon}^{-2} \bfm$
    \STATE $\hat{k}_{\eta}^{c}(\bfX^*, \bfX^*) :=  \bfS$
    \STATE \textbf{return} $\hat{\mu}_{\eta}^{c}(\bfX^*)$, $\hat{k}_{\eta}^{c}(\bfX^*, \bfX^*)$
\end{algorithmic}
\end{algorithm}

\section{Distributed GP with optimal weights}
In this section, we outline the algorithms of the proposed aggregation model for distributed SVGP and distributed exact GP in  \Cref{subsec:distributed svgp algs} and
\Cref{subsec:distributed exactgp algs} respectively.
\subsection{SVGP}
\label{subsec:distributed svgp algs}
\begin{algorithm}[H]
\caption{Optimal Weights for Distributed SVGP}
\label{alg:distributed svgp optimal weights}
\setstretch{0.99} 
\begin{algorithmic}[1]
    \STATE {\bfseries Input:} local datasets $\{\bfX_i, \bfy_i\}_{i=1}^{M}$, noise variance $\sigma_{\epsilon}^{2}$, a GP denoted by $\calGP( \mu(\cdot), k(\cdot,\cdot'))$, inducing inputs $\bfZ=\{\bfz_i\}_{i=1}^{m}$
    \STATE {\bfseries Output:} optimal weights $\bfbeta = \{ \beta_i \}_{i=1}^{M}$, kernel weights $\{ \bfalpha_{i} \}_{i=1}^{M}$, Cholesky factors $\{ \bfL_{i} \}_{i=1}^{M}$
    \STATE $\bfX_c := \{\bfx_i^{(c)}\}_{i=1}^{M}$, $\bfx_i^{(c)} \in \bfX_i$ is randomly selected
    \STATE $\bfM^{(c)}= \big[ k( \bfZ, \bfZ ) + \sigma_{\epsilon}^{-2} k( \bfZ, \bfX_c) k( \bfX_c, \bfZ ) \big]$
    \STATE $\bfA:=\text{zeros}(M,M)$ 
    \FOR[for loop over $t$ and $i$ can be parallelized]{$t=0$ {\bfseries to} $M-1$}
        \FOR[loop over the diagonal with offset $=t$]{$i=1$ {\bfseries to} $M-t$}
        \STATE $j := i+t$
         \IF[$i=j$]{$t = 0$ }
             \STATE $\bfM_{} = \big[ k( \bfZ, \bfZ ) + \sigma_{\epsilon}^{-2} k( \bfZ, \bfX_i) k( \bfX_i, \bfZ ) \big]$
            \STATE $\bfL_{i} := \text{chol}(\bfM)$ s.t. $\bfL_{i} \bfL_{i}^{\top} = \bfM$
            \STATE $\bfR :=\bfL_{i} \,\backslash\, \big[ k(\bfZ, \bfX_i) \bfy_i \big]$
            \STATE $\bfalpha_{i} := \bfL_{i}^{\top} \,\backslash\, \bfR $
        \ENDIF
        \STATE $\bfA[i,j] = \bfA[j,i] = \sigma_{\epsilon}^{-2} \bfalpha_{i}^{\top} \bfM^{(c)} \bfalpha_{j}$
        \ENDFOR
    \ENDFOR
    \STATE $\bfbeta := \bfA \,\backslash\, \text{diag}(\bfA)$
    \STATE \textbf{return} $\bfbeta$, $\{ \bfalpha_{i} \}_{i=1}^{M}$, $\{ \bfL_{i} \}_{i=1}^{M}$
\end{algorithmic}
\end{algorithm}

\begin{algorithm}[H]
\caption{Aggregated Prediction for Distributed SVGP}
\label{alg:distributed svgp aggreg pred}
\setstretch{0.99} 
\begin{algorithmic}[1]
    \STATE {\bfseries Input:} local datasets $\{\bfX_i, \bfy_i\}_{i=1}^{M}$, noise variance $\sigma_{\epsilon}^{2}$, test inputs $\bfX^* = \{ \bfx_i^* \}_{i=1}^{n_t}$, a GP denoted by $\calGP( \mu(\cdot), k(\cdot,\cdot'))$, inducing inputs $\bfZ=\{\bfz_i\}_{i=1}^{m}$
    \STATE {\bfseries Output:} joint mean $\mu_{\calA}(\bfX^*)$ and joint variance $\sigma_{\calA}^{2}(\bfX^*)$ in \cref{eq:aggreg pred svgp}
    \STATE get $\bfbeta = ( \beta_1, \ldots, \beta_M )^{\top}$, $\{ \bfalpha_{i} \}_{i=1}^{M}$, $\{ \bfL_{i} \}_{i=1}^{M}$ from \Cref{alg:distributed svgp optimal weights}
    \STATE $\bfm := \text{zeros}(n_t)$; $\bfs := \text{zeros}(n_t)$
    \FOR[for loop over $i$ can be parallelized]{$i=1$ {\bfseries to} $M$}
    \STATE $\bfm \leftarrow \bfm + \beta_{i} k(\bfX^*, \bfZ) \bfalpha_{i}$
    \STATE $ \bfR := \bfL_i \backslash K(\bfZ, \bfX^*)$
    \STATE $\bfs \leftarrow \bfs + \beta_i^2  \bfR^{\top} \bfR$
    \ENDFOR
    \STATE $\mu_{\calA}(\bfX^*) := \sigma_{\epsilon}^{-2} \bfm$
    \STATE $\sigma_{\calA}^{2}(\bfX^*) :=  \bfs$
    \STATE \textbf{return} $\mu_{\calA}(\bfX^*)$, $\sigma_{\calA}^{2}(\bfX^*)$
\end{algorithmic}
\end{algorithm}

\subsection{Exact GP}
\label{subsec:distributed exactgp algs}
\begin{algorithm}[H]
\caption{Optimal Weights for Distributed Exact GP}
\label{alg:distributed exactgp optimal weights}
\setstretch{0.99} 
\begin{algorithmic}[1]
    \STATE {\bfseries Input:} local datasets $\{\bfX_i, \bfy_i\}_{i=1}^{M}$, noise variance $\sigma_{\epsilon}^{2}$, a GP denoted by $\calGP( \mu(\cdot), k(\cdot,\cdot'))$
    \STATE {\bfseries Output:} optimal weights $\bfbeta = \{ \beta_i \}_{i=1}^{M}$, kernel weights $\{ \bfalpha_{i} \}_{i=1}^{M}$, Cholesky factors $\{ \bfL_{i} \}_{i=1}^{M}$
    \STATE $\bfX_c := \{\bfx_i^{(c)}\}_{i=1}^{M}$, $\bfx_i^{(c)} \in \bfX_i$ is randomly selected
    \STATE $\bfA:=\text{zeros}(M,M)$ 
    \FOR[for loop over $t$ and $i$ can be parallelized]{$t=0$ {\bfseries to} $M-1$}
        \FOR[loop over the diagonal with offset $=t$]{$i=1$ {\bfseries to} $M-t$}
        \STATE $j := i+t$
        \STATE $\bfM_{}^{(c)} = \big[ k( \bfX_i, \bfX_j ) + k( \bfX_i, \bfX_c) k( \bfX_c, \bfX_i ) \big]$
         \IF[$i=j$]{$t = 0$ }
            \STATE $\bfL_{i} := \text{chol}(\widetilde{\bfK}_{\bff_i \bff_i})$ s.t. $\bfL_{i} \bfL_{i}^{\top} = \widetilde{\bfK}_{\bff_i \bff_i} = k(\bfX_i, \bfX_i) + \sigma_{\epsilon}^2 \bfI_{n_i}$
            \STATE $\bfR :=\bfL_{i} \,\backslash\, \bfy_i$
            \STATE $\bfalpha_{i} := \bfL_{i}^{\top} \,\backslash\, \bfR $
        \ENDIF
        \STATE $\bfA[i,j] = \bfA[j,i] = \bfalpha_{i}^{\top} \bfM^{(c)} \bfalpha_{j}$
        \ENDFOR
    \ENDFOR
    \STATE $\bfbeta := \bfA \,\backslash\, \text{diag}(\bfA)$
    \STATE \textbf{return} $\bfbeta$, $\{ \bfalpha_{i} \}_{i=1}^{M}$, $\{ \bfL_{i} \}_{i=1}^{M}$
\end{algorithmic}
\end{algorithm}

\begin{algorithm}[H]
\caption{Aggregated Prediction for Distributed Exact GP}
\label{alg:distributed exactgp aggreg pred}
\setstretch{0.99} 
\begin{algorithmic}[1]
    \STATE {\bfseries Input:} local datasets $\{\bfX_i, \bfy_i\}_{i=1}^{M}$, noise variance $\sigma_{\epsilon}^{2}$, test inputs $\bfX^* = \{ \bfx_i^* \}_{i=1}^{n_t}$, a GP denoted by $\calGP( \mu(\cdot), k(\cdot,\cdot'))$
    \STATE {\bfseries Output:} joint mean $\mu_{\calA}(\bfX^*)$ and joint variance $\sigma_{\calA}^{2}(\bfX^*)$ in \cref{eq:aggreg pred exactgp}
    \STATE get $\bfbeta = ( \beta_1, \ldots, \beta_M )^{\top}$, $\{ \bfalpha_{i} \}_{i=1}^{M}$, $\{ \bfL_{i} \}_{i=1}^{M}$ from \Cref{alg:distributed exactgp optimal weights}
    \STATE $\bfm := \text{zeros}(n_t)$; $\bfs := \text{zeros}(n_t)$
    \FOR[for loop over $i$ can be parallelized]{$i=1$ {\bfseries to} $M$}
    \STATE $\bfm \leftarrow \bfm + \beta_{i} k(\bfX^*, \bfX_i) \bfalpha_{i}$
    \STATE  $\bfR:= \bfL_i \backslash k(\bfX_i, \bfX^*)$
    \STATE $\bfs \leftarrow \bfs + \beta_i^2   \bfR^{\top} \bfR$
    \ENDFOR
    \STATE $\mu_{\calA}(\bfX^*) := \bfm$
    \STATE $\sigma_{\calA}^{2}(\bfX^*) :=  \bfs$
    \STATE \textbf{return} $\mu_{\calA}(\bfX^*)$, $\sigma_{\calA}^{2}(\bfX^*)$
\end{algorithmic}
\end{algorithm}

\section{Experiment}
In this section, we provide the supporting boxplots for the distributed training on synthetic data and tables for the aggregation models on UCI datasets in \Cref{subsec:boxplots hyper estimate}.
\subsection{Boxplots of the hyperparameter estimates}
\label{subsec:boxplots hyper estimate}
\begin{figure}[H]
    \centering
    \begin{minipage}{\textwidth}
        \centering
        \includegraphics
        [height=0.105\textheight]{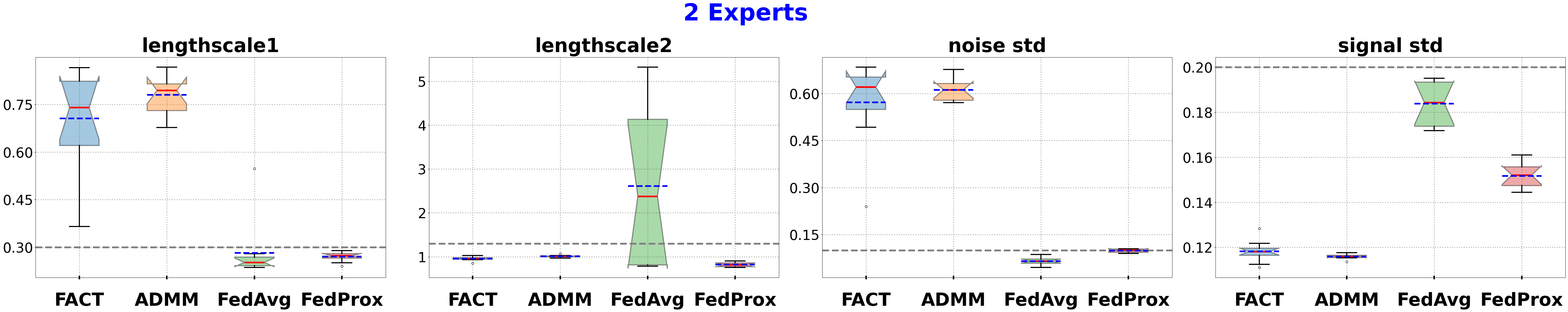}
    \end{minipage}
    \vfill
    \begin{minipage}{\textwidth}
        \centering
        \includegraphics
        [height=0.105\textheight]{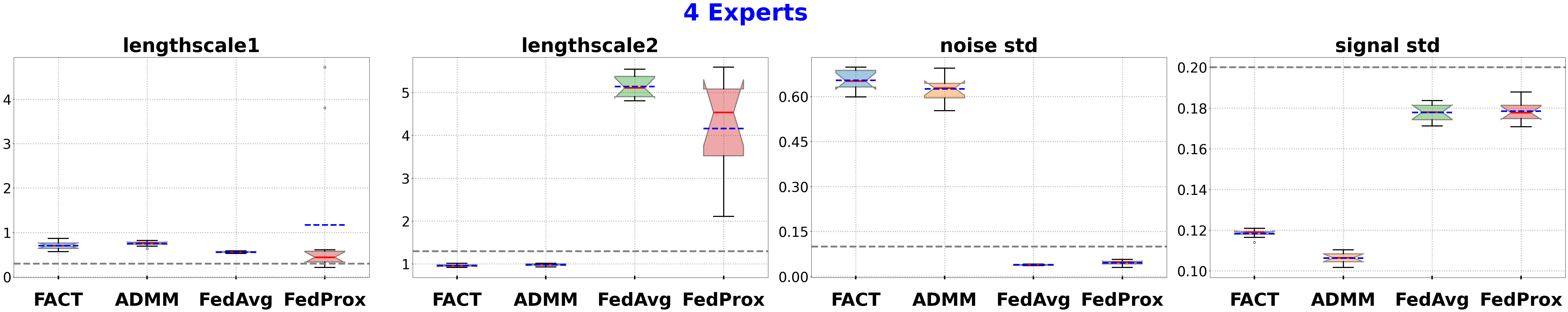}
    \end{minipage}
    \vfill
    \begin{minipage}{\textwidth}
        \centering
        \includegraphics
        [height=0.105\textheight]{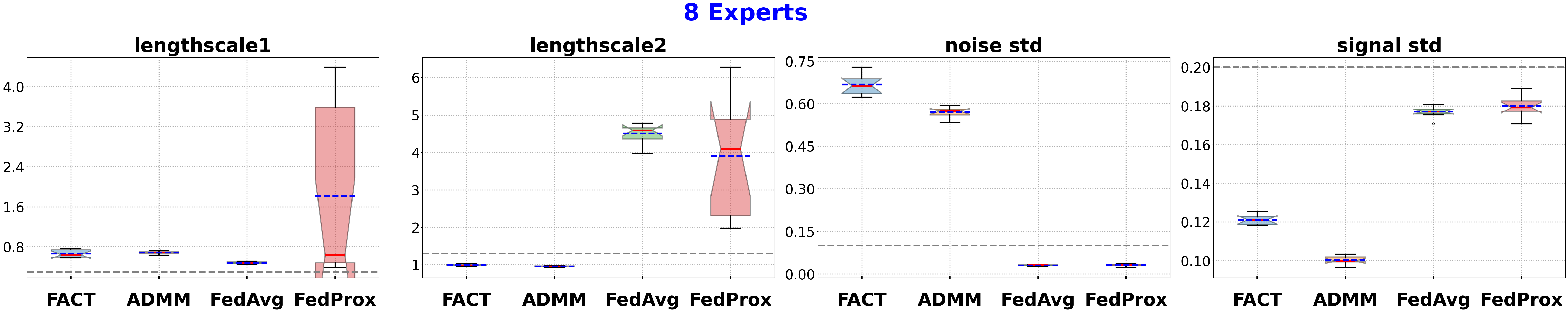}
    \end{minipage}
    \vfill
    \begin{minipage}{\textwidth}
        \centering
        \includegraphics
        [height=0.105\textheight]{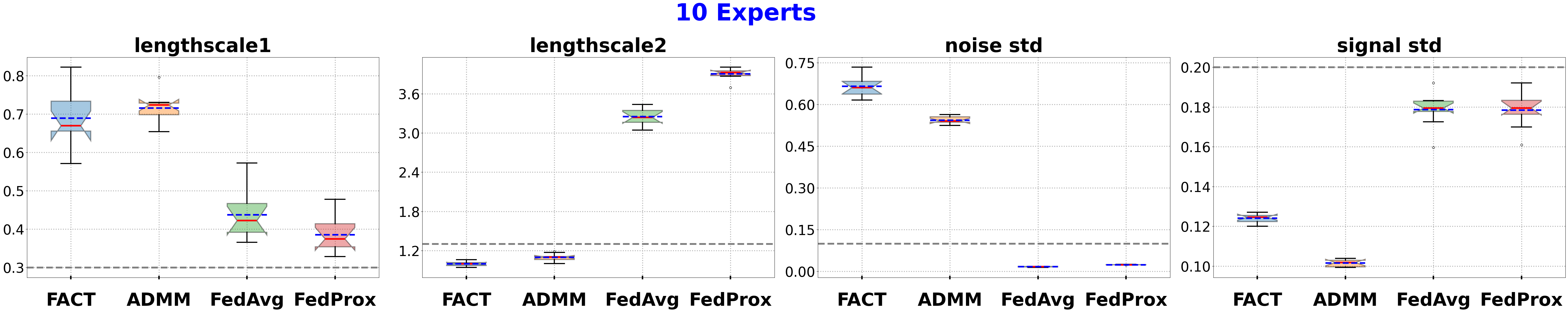}
    \end{minipage}
    \vfill
    \begin{minipage}{\textwidth}
        \centering
        \includegraphics
        [height=0.105\textheight]{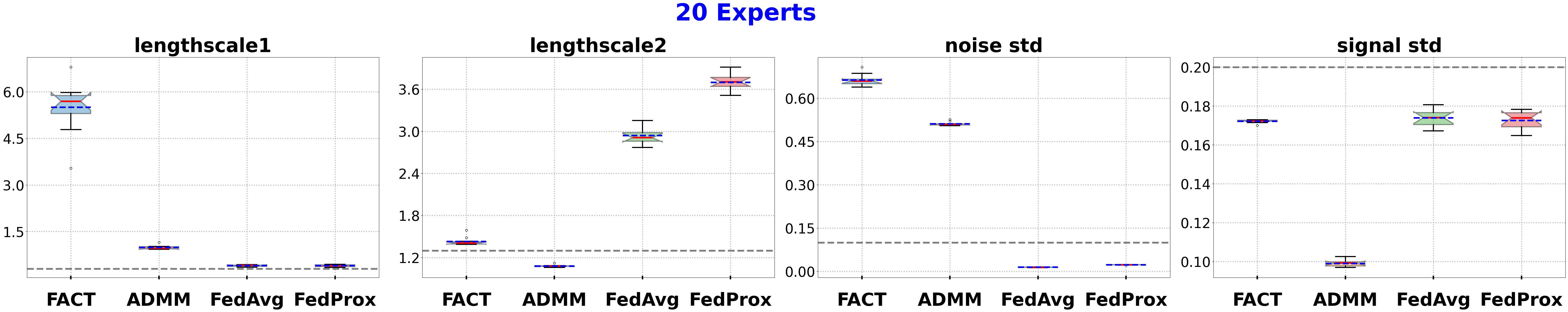}
    \end{minipage}
    \vfill
    \begin{minipage}{\textwidth}
        \centering
        \includegraphics
        [height=0.105\textheight]{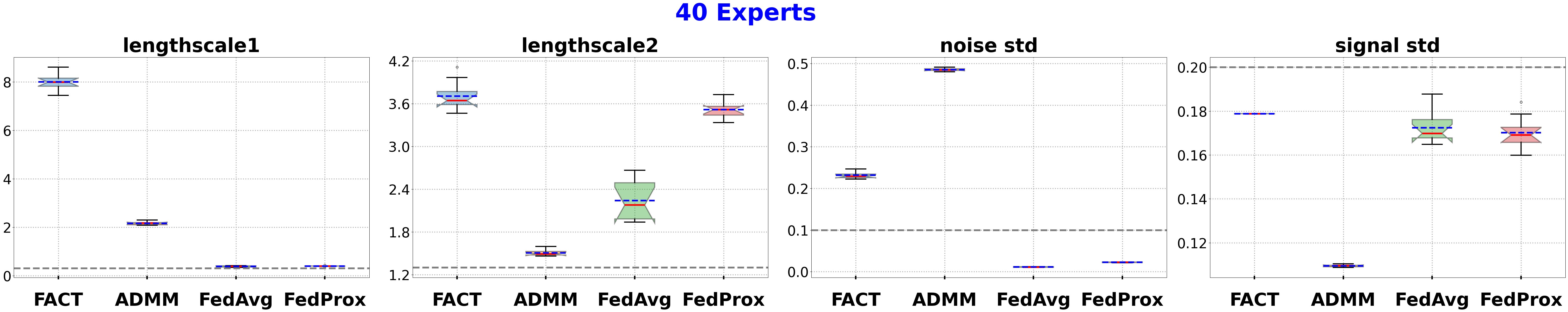}
    \end{minipage}
    \vfill
    \begin{minipage}{\textwidth}
        \centering
        \includegraphics
        [height=0.105\textheight]{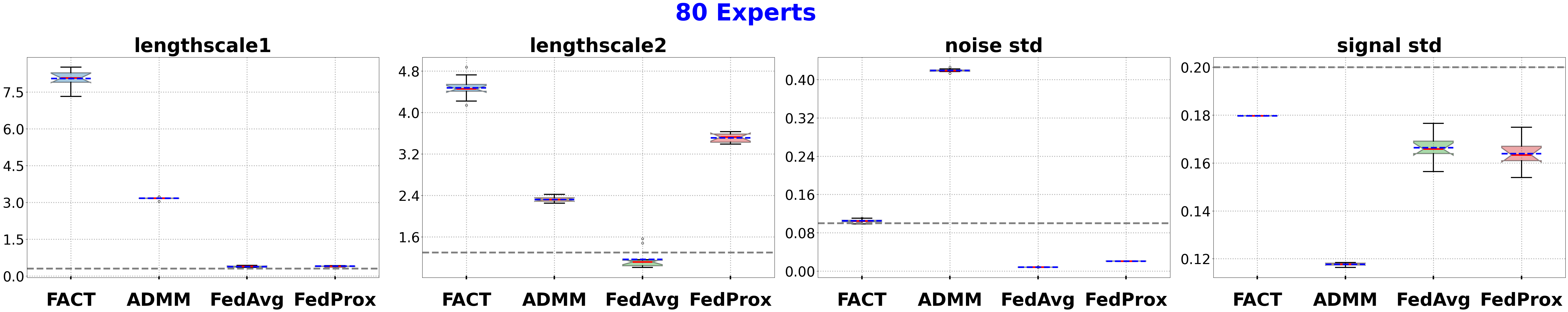}
    \end{minipage}
    \vfill
    \begin{minipage}{\textwidth}
        \centering
        \includegraphics
        [height=0.105\textheight]{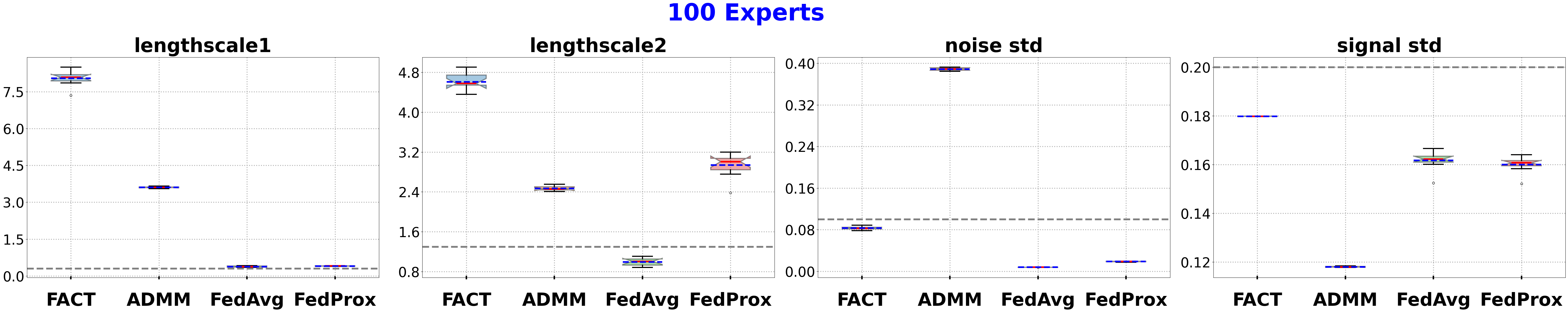}
    \end{minipage}
    \caption{Boxplots of the optimized hyperparameter estimates on the $n=10^4$ dataset.}
    \label{fig:train_hyperparam_boxplots}
\end{figure}

\end{document}